\documentclass[runningheads]{llncs}

\usepackage{eccv}

\usepackage{eccvabbrv}

\usepackage{graphicx}
\usepackage{booktabs}
\usepackage{float}
\usepackage{siunitx}
\usepackage[export]{adjustbox}
\usepackage{tabularx}
\usepackage[table]{xcolor}
\usepackage{multirow}
\usepackage{amssymb} %
\newcommand{\cmark}{\checkmark}
\usepackage{tikz}
\usetikzlibrary{shapes.geometric, shapes, arrows, arrows.meta, bending}
\usepackage{pgfplots}
\usepgfplotslibrary{groupplots}
\pgfplotsset{compat=1.18}

\usepackage[accsupp]{axessibility}  %

\usepackage{hyperref}

\hypersetup{
  pdftitle={MAGneT-3D: Monocular And Domain-Generalizable Temporal 3D Detection},
  pdfauthor={Mohamed Kotb, Johannes Meier, Christoph Reich, Oussema Dhaouadi, Luis Denninger, Daniel Cremers},
  pdfkeywords={Monocular 3D Object Detection, Domain Generalization, Temporal Modeling},
}

\usepackage{orcidlink}

\newcommand{\anchorgen}{Domain-Robust Anchor Generator}
\newcommand{\anchorgensymbol}{DRAG\xspace}
\newcommand{\propgationtransformer}{Temporal Refinement and Identity Merging}
\newcommand{\propgationtransformersymbol}{TRIM\xspace}

\newcommand{\methodsymbol}{MAGneT-3D\xspace}

\newcommand \colorindicator[1]{%
	{\textcolor{#1}{$\blacksquare\!\!\!\!\!\blacksquare$}}%
}

\definecolor{colorgt}{RGB}{0,128,0}
\definecolor{colorours}{RGB}{0,0,255}
\definecolor{colorbaseline}{RGB}{255,165,0}

\definecolor{tud0d}{RGB}{83,83,83}
\definecolor{tud0c}{RGB}{137,137,137}
\definecolor{tud0b}{RGB}{181,181,181}
\definecolor{tud0a}{RGB}{220,220,220}
\definecolor{tud1a}{RGB}{93,133,195}
\definecolor{tud2a}{RGB}{0,156,218}
\definecolor{tud3a}{RGB}{80,182,149}
\definecolor{tud4a}{RGB}{175,204,80}
\definecolor{tud5a}{RGB}{221,223,72}
\definecolor{tud6a}{RGB}{255,224,92}
\definecolor{tud7a}{RGB}{248,186,60}
\definecolor{tud8a}{RGB}{238,122,52}
\definecolor{tud9a}{RGB}{233,80,62}
\definecolor{tud10a}{RGB}{201,48,142}
\definecolor{tud11a}{RGB}{128,69,151}
\definecolor{tud1b}{RGB}{0,90,169}
\definecolor{tud2b}{RGB}{0,131,204}
\definecolor{tud3b}{RGB}{0,157,129}
\definecolor{tud4b}{RGB}{153,192,0}
\definecolor{tud5b}{RGB}{201,212,0}
\definecolor{tud6b}{RGB}{253,202,0}
\definecolor{tud7b}{RGB}{245,163,0}
\definecolor{tud8b}{RGB}{236,101,0}
\definecolor{tud9b}{RGB}{230,0,26}
\definecolor{tud10b}{RGB}{166,0,132}
\definecolor{tud11b}{RGB}{114,16,133}
\definecolor{tud1c}{RGB}{0,78,138}
\definecolor{tud2c}{RGB}{0,104,157}
\definecolor{tud3c}{RGB}{0,136,119}
\definecolor{tud4c}{RGB}{127,171,22}
\definecolor{tud5c}{RGB}{177,189,0}
\definecolor{tud6c}{RGB}{215,172,0}
\definecolor{tud7c}{RGB}{210,135,0}
\definecolor{tud8c}{RGB}{204,76,3}
\definecolor{tud9c}{RGB}{185,15,34}
\definecolor{tud10c}{RGB}{149,17,105}
\definecolor{tud11c}{RGB}{97,28,115}
\definecolor{tud1d}{RGB}{36,53,114}
\definecolor{tud2d}{RGB}{0,78,115}
\definecolor{tud3d}{RGB}{0,113,94}
\definecolor{tud4d}{RGB}{106,139,55}
\definecolor{tud5d}{RGB}{153,166,4}
\definecolor{tud6d}{RGB}{174,142,0}
\definecolor{tud7d}{RGB}{190,111,0}
\definecolor{tud8d}{RGB}{169,73,19}
\definecolor{tud9d}{RGB}{156,28,38}
\definecolor{tud10d}{RGB}{115,32,84}
\definecolor{tud11d}{RGB}{76,34,106}

\definecolor{sfov}{RGB}{0,128,0}
\definecolor{tfov}{RGB}{255,100,255}
\definecolor{fovgap}{RGB}{255,100,255}

\definecolor{c1}{RGB}{238, 204, 102}
\definecolor{c2}{RGB}{238, 153, 170}
\definecolor{c3}{RGB}{102, 153, 204}
\definecolor{c4}{RGB}{153, 119, 0}
\definecolor{c5}{RGB}{153, 68, 85}

\definecolor{ours1}{RGB}{216,228,230}
\definecolor{ours2}{RGB}{8,48,107}

\definecolor{spetr1}{RGB}{250,230,200}
\definecolor{spetr2}{RGB}{127,39,4}

\definecolor{gt1}{RGB}{210,230,180}
\definecolor{gt2}{RGB}{0,68,27}

\begin{document}

\title{MAGneT-3D: Monocular and Domain-Generalizable Temporal 3D Detection}

\titlerunning{Monocular \& Domain-Generalizable Temporal 3D Detection}

\author{Mohamed Kotb\inst{1,\dag\,}\orcidlink{0000-0002-8511-7149} \and
Johannes Meier\inst{1,3,5,\dag\,}\orcidlink{0009-0000-2227-8271} \and
Christoph Reich\inst{1,2,5,6\,}\orcidlink{0000-0002-8616-1627} \and\\
Oussema Dhaouadi\inst{1,3,4,5,7\,}\orcidlink{0009-0008-6842-5220} \and
Luis Denninger\inst{1,5,7\,}\orcidlink{0009-0001-2913-7938} \and
Daniel Cremers\inst{1,5,6\,}\orcidlink{0000-0002-3079-7984}}

\authorrunning{M. Kotb, J. Meier \emph{et al.}}

\institute{
\textsuperscript{1}\,TU Munich\;\;\;\;\;
\textsuperscript{2}\,TU Darmstadt\;\;\;\;\;
\textsuperscript{3}\,ETH Zurich\;\;\;\;\;
\textsuperscript{4}\,University of Cambridge\\
\textsuperscript{5}\,MCML\;\;\;\;\;
\textsuperscript{6}\,ELIZA\;\;\;\;\;
\textsuperscript{7}\,DeepScenario\;\;\;\;\;\textsuperscript{\dag}\,Equal contribution\\
\email{\{mohamed.kotb,\,j.meier\}@tum.de}\\
\url{https://mo-sameh.github.io/MAGneT-3D-Project-Page/}
}

\maketitle

\begin{abstract}
Monocular temporal 3D detection aims to detect objects in 3D, given a monocular video. Query-based 3D detectors unify detection and cross-view association, but their learnable queries fit the spatial distribution of the training data (\eg, field-of-view). We show that this issue is especially severe when these models are applied to monocular video, hindering generalization to unseen datasets and environments. To address this limitation, we introduce MAGneT-3D, the first method for domain-generalized monocular temporal 3D object detection. Instead of relying on static learnable queries, we propose a Domain-Robust Anchor Generator (DRAG) approach that adaptively derives 3D proposals during inference. To further enable domain generalization, we propose a Temporal Refinement and Identity Merging (TRIM) strategy, reducing dependence on specific 3D proposals. To enable comprehensive domain-generalization evaluation, we establish a cross-dataset benchmark spanning nuScenes, Waymo, Lyft, and ONCE. Under zero-shot domain shifts, MAGneT-3D outperforms all baselines, improving NDS from \SI{12.1}{\%} to \SI{18.6}{\%} while also increasing in-domain accuracy. %
\end{abstract}

\section{Introduction}
\label{sec:intro}

Accurate 3D perception forms a cornerstone for autonomous driving \cite{meier2025lead,monolss,monotakd}, robotics\cite{robotic1,robotic2}, and intelligent surveillance \cite{rope3d,carladrone,ideal-m3d}. Monocular 3D detection aims to detect and classify objects in 3D from a single image. While the monocular setting enables cost-efficient deployment of such detectors, monocular approaches struggle with depth ambiguities \cite{monocd,meier2025lead}, limiting their accuracy and generalization. LiDAR-based detectors can overcome these limitations but require specialized hardware, which significantly constrains applicability \cite{3dssd,Std:Sparse-to-dense,dbqssd}. Instead, we leverage multi-frame monocular cues for 3D detection to overcome the limitations of single-frame detectors, maximizing applicability.

While recent work has proposed various monocular 3D detectors \cite{meier2025lead,mondiff,monoatt,mogde,monouni}, these typically do not take advantage of multi-frame cues. Some monocular detectors use heuristic post-processing or simple recurrent architectures to encode multi-frame information, but do not leverage current transformer-based architectures. Other approaches exploit synchronization multi-view data for 3D detection \cite{streampetr,sparse4dv3,far3d}. These approaches utilize current transformer-based detectors with query-based decoders to model multi-view context and enhance detection accuracy \cite{streampetr,sparse4dv3,far3d}. However, as we show, query-based decoders entail significant biases introduced by the training data. While learnable, queries are static during inference and fit to the spatial distribution (\eg, field-of-view) of the training data. This is especially prominent when multi-view models are trained using monocular video, and significantly hampers domain generalization.

\begin{figure}[t]
    \centering
    \begin{minipage}{.498\textwidth}
        \centering
            \begin{tikzpicture}[
        every node/.style={align=center, font=\sffamily\tiny},
        scale=0.65,
        onlytext/.style={
            font=\sffamily\tiny,
            text width=1.65cm,
            inner sep=1pt
        },
        intertext/.style={
            font=\sffamily\tiny,
            text width=1.75cm,
            inner sep=1pt
        }
    ]

        \def\r{2.5}

        \coordinate (A) at (-1.2,  1.0);
        \coordinate (B) at ( 1.2,  1.0);
        \coordinate (C) at ( 0.0, -1.0);

        \fill[c3!80,  opacity=0.35] (A) circle (\r);
        \fill[c1!80, opacity=0.35] (B) circle (\r);
        \fill[c5!80,   opacity=0.35] (C) circle (\r);

        \draw[thick, black] (A) circle (\r);
        \draw[thick, black] (B) circle (\r);
        \draw[thick, black] (C) circle (\r);

        \node[font=\sffamily\small] at (-3.55, 3.45) {\textbf{Monocular}};
        \node[font=\sffamily\small] at ( 3.55, 3.45) {\textbf{Temporal}};
        \node[font=\sffamily\small] at ( 0.00,-3.9) {\textbf{Domain Generalization}};

        \node[onlytext] at (-1.50, 2.75) {MonoCD~\cite{monocd}};
        \node[onlytext] at (-2.20, 2.00) {FCOS3D~\cite{fcos3d}};
        \node[onlytext] at (-2.50, 1.25) {LeAD-M3D~\cite{meier2025lead}};
        \node[onlytext] at (-2.87, 0.50) {SMOKE~\cite{smoke}};

        \node[onlytext] at (1.75, 2.75) {StreamPETR~\cite{streampetr}};
        \node[onlytext] at ( 2.275, 2.00) {Sparse4D-V3~\cite{sparse4dv3}};
        \node[onlytext] at ( 2.50, 1.25) {BEVFormer V2~\cite{bevformerv2}};

        \node[onlytext] at (-1.15,-2.0) {3D-VField~\cite{3dvfield}};
        \node[onlytext] at ( 0.25,-2.75) {DG-BEV~\cite{dgbev}};
        \node[onlytext] at ( 1.40,-2.0) {LSF~\cite{lsf}};

        \node[intertext] at ( 0.00, 2.00)
            {MonoTDF~\cite{monotdf}};

        \node[intertext] at (-1.45,-1.05)
            {MonoGDG~\cite{monodgp}};

        \node[intertext] at ( 1.4,-1.05)
            {TSMA-BEV~\cite{tsmabev}};

        \node[
            intertext,
            font=\sffamily\scriptsize
        ] at (0.00,0.30)
            {\textbf{MAGneT-3D}\\\textbf{(Ours)}};

    \end{tikzpicture}%
    \end{minipage}%
    \hfill%
    \begin{minipage}{0.498\textwidth}
        \centering
            \begin{tikzpicture}[
        every node/.style={font=\sffamily\tiny},
        scale=0.715,
        grid/.style={
            draw=gray!30,
            line width=0.45pt,
            dashed,
        },
        axis/.style={
            draw=gray!45,
            line width=0.5pt
        },
        frame/.style={
            draw=black,
            line width=0.9pt
        },
        ticklabel/.style={
            font=\scriptsize
        },
        axislabel/.style={
            font=\small
        },
        legendlabel/.style={
            font=\scriptsize
        },
        legendline/.style={
            line width=1.2pt
        }
    ]

    \pgfmathsetmacro{\Rframe}{2.55}

    \pgfmathsetmacro{\Rgrid}{2.25}

    \pgfmathsetmacro{\RingStep}{\Rgrid/6}

    \pgfmathsetmacro{\Rtwo}{2*\RingStep}
    \pgfmathsetmacro{\Rfour}{4*\RingStep}
    \pgfmathsetmacro{\Rsix}{6*\RingStep}

    \pgfmathsetmacro{\TickInset}{0}

    \pgfmathsetmacro{\TopMin}{10}
    \pgfmathsetmacro{\TopMid}{20}
    \pgfmathsetmacro{\TopMax}{30}

    \pgfmathsetmacro{\SideMin}{7}
    \pgfmathsetmacro{\SideMid}{15}
    \pgfmathsetmacro{\SideMax}{22}

    \def\TopAxisTitle{\sffamily\small \textcolor{gray}{\textbf{Source-nuScenes}}}
    \def\RightAxisTitle{\sffamily\small \textbf{Target}\\\sffamily\small \textbf{ONCE}}
    \def\BottomAxisTitle{\sffamily\small \textbf{Target-Lyft}}
    \def\LeftAxisTitle{\sffamily\small \textbf{Target}\\\sffamily\small \textbf{Waymo}}

    \def\SPTop{28.5}
    \def\SPRight{12.7}
    \def\SPBottom{10.6}
    \def\SPLeft{13.1}

    \def\SVTop{27.0}
    \def\SVRight{14.7}
    \def\SVBottom{13.1}
    \def\SVLeft{9.5}

    \def\FTop{29.8}
    \def\FRight{13.2}
    \def\FBottom{12.8}
    \def\FLeft{10.8}

    \def\BTop{22.2}
    \def\BRight{14.5}
    \def\BBottom{9.1}
    \def\BLeft{11.2}

    \def\MTop{32.9}
    \def\MRight{19.1}
    \def\MBottom{19.5}
    \def\MLeft{17.2}

    \newcommand{\GetTopRadius}[2]{%
        \pgfmathsetmacro{#2}{%
            \Rtwo
            + (\Rsix-\Rtwo)
            * ((#1-\TopMin)/(\TopMax-\TopMin))
        }%
    }

    \newcommand{\GetSideRadius}[2]{%
        \pgfmathparse{%
            #1 <= \SideMid
            ?
            \Rtwo
            + (\Rfour-\Rtwo)
            * ((#1-\SideMin)/(\SideMid-\SideMin))
            :
            \Rfour
            + (\Rsix-\Rfour)
            * ((#1-\SideMid)/(\SideMax-\SideMid))
        }%
        \edef#2{\pgfmathresult}%
    }

    \newcommand{\RadarSeries}[7]{%
        \GetTopRadius{#2}{\RadiusTop}
        \GetSideRadius{#3}{\RadiusRight}
        \GetSideRadius{#4}{\RadiusBottom}
        \GetSideRadius{#5}{\RadiusLeft}

        \draw[
            draw=#1,
            draw opacity=1,
            line width=#6,
            line join=round,
        ]
            ( 90:\RadiusTop)
            --
            (  0:\RadiusRight)
            --
            (270:\RadiusBottom)
            --
            (180:\RadiusLeft)
            -- cycle;
        \node[fill=none] at (90:\RadiusTop) {\textcolor{#1}{\normalsize $\star$}};
        \node[fill=none] at (0:\RadiusRight) {\textcolor{#1}{\normalsize $\star$}};
        \node[fill=none] at (270:\RadiusBottom) {\textcolor{#1}{\normalsize $\star$}};
        \node[fill=none] at (180:\RadiusLeft) {\textcolor{#1}{\normalsize $\star$}};
    }

    \foreach \i in {1,...,6}{
        \draw[grid]
            (0,0) circle ({\i*\RingStep});
    }

    \draw[frame]
        (0,0) circle (\Rframe);

    \draw[axis] (0,-\Rsix) -- (0,\Rsix);
    \draw[axis] (-\Rsix,0) -- (\Rsix,0);

    \RadarSeries
        {c3}
        {\SPTop}{\SPRight}{\SPBottom}{\SPLeft}
        {0.6pt}{0.11}

    \RadarSeries
        {c4}
        {\SVTop}{\SVRight}{\SVBottom}{\SVLeft}
        {0.6pt}{0.11}

    \RadarSeries
        {c2}
        {\FTop}{\FRight}{\FBottom}{\FLeft}
        {0.6pt}{0.11}

    \RadarSeries
        {c5}
        {\BTop}{\BRight}{\BBottom}{\BLeft}
        {0.6pt}{0.11}

    \RadarSeries
        {c1}
        {\MTop}{\MRight}{\MBottom}{\MLeft}
        {0.9pt}{0.20}

    \node[axislabel]
        at (0,\Rframe+0.28)
        {\TopAxisTitle};

    \node[axislabel, align=center]
        at (\Rframe+0.7,0)
        {\RightAxisTitle};

    \node[axislabel]
        at (0,-\Rframe-0.33)
        {\BottomAxisTitle};

    \node[axislabel, align=center]
        at (-\Rframe-0.75,0)
        {\LeftAxisTitle};

    \begin{scope}[xshift=0.8cm]
    \draw[legendline, c3]
        (-4.25,-3.35) -- (-3.75,-3.35);

    \node[legendlabel, anchor=west]
        at (-3.62,-3.35)
        {\sffamily\tiny StreamPETR~\cite{streampetr} (Baseline)};

     \begin{scope}[xshift=0.2cm]
    \draw[legendline, c2]
        (-0.45,-3.35) -- (0.05,-3.35);

    \node[legendlabel, anchor=west]
        at (0.18,-3.35)
        {\sffamily\tiny Far3D~\cite{far3d}};
    \end{scope}

    \draw[legendline, c1]
        (-4.25,-4.01) -- (-3.75,-4.01);

    \node[legendlabel, anchor=west]
        at (-3.62,-4.01)
        {\sffamily\tiny MAGneT-3D (Ours)};

    \draw[legendline, c4]
        (-4.25,-3.68) -- (-3.75,-3.68);

    \node[legendlabel, anchor=west]
        at (-3.62,-3.68)
        {\sffamily\tiny Sparse4D v3~\cite{sparse4dv3}};

    \begin{scope}[xshift=0.2cm]
    \draw[legendline, c5]
        (-0.45,-3.68) -- (0.05,-3.68);

    \node[legendlabel, anchor=west]
        at (0.18,-3.68)
        {\sffamily\tiny BEVFormer v2~\cite{bevformerv2}};
        \end{scope}
    \end{scope}

    \node[ticklabel, anchor=center]
        at (0,\Rtwo-\TickInset) {\sffamily\small 10};

    \node[ticklabel, anchor=center]
        at (0,\Rfour-\TickInset) {\sffamily\small 20};

    \node[ticklabel, anchor=center]
        at (0,\Rsix-\TickInset) {\sffamily\small 30};

    \node[ticklabel, anchor=center]
        at (\Rtwo-\TickInset,0) {\sffamily\small 7};

    \node[ticklabel, anchor=center]
        at (\Rfour-\TickInset,0) {\sffamily\small 15};

    \node[ticklabel, anchor=center]
        at (\Rsix-\TickInset,0) {\sffamily\small 22};

    \node[ticklabel, anchor=center]
        at (0,-\Rtwo+\TickInset) {\sffamily\small 7};

    \node[ticklabel, anchor=center]
        at (0,-\Rfour+\TickInset) {\sffamily\small 15};

    \node[ticklabel, anchor=center]
        at (0,-\Rsix+\TickInset) {\sffamily\small 22};

    \node[ticklabel, anchor=center]
        at (-\Rtwo+\TickInset,0) {\sffamily\small 7};

    \node[ticklabel, anchor=center]
        at (-\Rfour+\TickInset,0) {\sffamily\small 15};

    \node[ticklabel, anchor=center]
        at (-\Rsix+\TickInset,0) {\sffamily\small 22};

\end{tikzpicture}%
    \end{minipage}
    \vspace{-2pt}
    \caption{\textbf{\methodsymbol{} overview.} \methodsymbol{} bridges monocular, temporal modeling and domain generalization \emph{(left)} for strong cross-dataset 3D detection \emph{(right)}. Prior detectors instead treat these settings as distinct or paired, leaving the intersection unexplored \emph{(left)}. On our new cross-dataset benchmark \emph{(right)}, \methodsymbol{} trained on nuScenes outperforms on both in-domain (source) and all unseen (target) datasets. We report the nuScenes Detection Score (NDS, in \%, $\uparrow$).}
    \label{fig:teaser}
\end{figure}
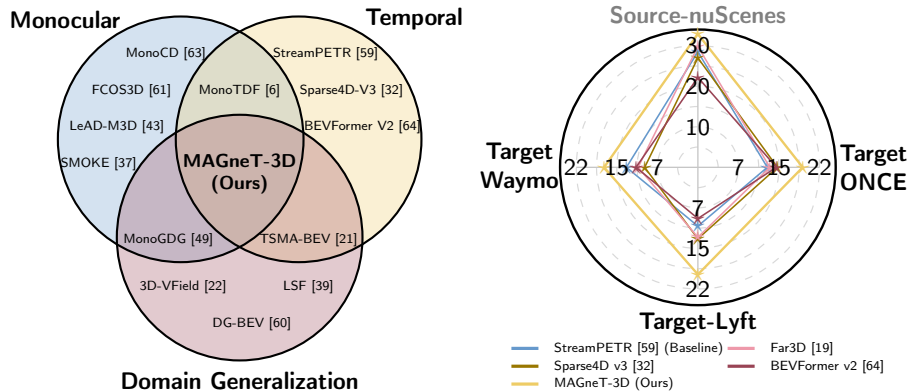

In this work, we approach monocular temporal 3D detection under domain shifts (\cf \cref{fig:teaser} \emph{(left)}). In particular, we propose \textbf{M}onocular \textbf{A}nd Domain-\textbf{G}e\textbf{ne}ralizable \textbf{T}emporal \textbf{3}D \textbf{D}etection (\textbf{MAGneT-3D}). To the best of our knowledge, MAGneT-3D is the first approach to explicitly address domain shifts for temporal monocular 3D detection. We overcome the limitations of static, learnable queries by introducing a Domain-Robust Anchor Generator (DRAG). This approach adaptively generates 3D proposals during inference based on image features. Additionally, we present a Temporal Refinement and Identity Merging (TRIM) strategy. This strategy pairs dense soft assignment with supervised contrastive learning to enhance detection diversity during training. For inference, TRIM uses a clustering approach to deduplicate predictions, improving precision. To enable a thorough and comprehensive domain generalization evaluation, we propose a novel cross-dataset benchmark, including four existing datasets. While the accuracy of existing approaches significantly suffers under domain shift, MAGneT-3D outperforms in domain generalization and also improves for in-domain data (\cf \cref{fig:teaser} \emph{(right)}). %

In particular, we make the following contributions: \emph{(i)} \textbf{Domain-Robust Anchor Generator (DRAG)}: We replace static, learnable detection queries with a dynamic anchor generator. By generating que\-ries on the fly from image features in high-confidence 3D regions, we eliminate source spatial bias and naturally adapt to varying camera configurations. \emph{(ii)} \textbf{\propgationtransformer{} (\propgationtransformersymbol{})}: During training, the proximity of our dynamic queries to the targets causes the refinement head to only learn minor, trivial adjustments. We address this by replacing the strict 1-to-1 Hungarian matching with a dense soft assignment strategy. To consolidate the resulting redundant predictions at inference, we introduce a clustering-based deduplication step. \emph{(iii)} \textbf{Temporal Domain Generalization Benchmark}: We propose a novel domain-generalization benchmark spanning four temporal 3D detection data\-sets: nuScenes \cite{nuscene}, Waymo \cite{waymo}, Lyft \cite{lyft}, and ONCE \cite{once}. MAGneT-3D scores \SI{18.6}{\%} in NDS average cross-dataset accuracy outperforming existing approaches under domain shifts, including our baseline StreamPETR \cite{streampetr} (\SI{12.1}{\%} in NDS). In-domain, MAGneT-3D achieves an NDS of \SI{32.9}{\%} on nuScenes \cite{nuscene}, surpassing existing approaches (\cf \cref{fig:teaser} \emph{(right)}).

\section{Related Work}
\label{sec:related_work}

\subsubsection{3D Object Detection.} Sensing modality strongly shapes both the robustness and cost of a 3D detector. LiDAR-based approaches achieve exceptional geometric accuracy by operating directly on point clouds \cite{3dssd,Std:Sparse-to-dense,dbqssd,Voxelnext,voxelrcnn,mssvt}. Alternatively, radar sensors provide complementary Doppler measurements with high resilience to adverse weather conditions \cite{centerfusion,transcar,rcbevdet}, while production-grade pipelines frequently fuse multiple modalities to maximize redundancy \cite{bevfusion,deepfusion,transfusion}. Due to the high cost of active sensors, camera-only perception is an effective alternative.

To mitigate the depth ambiguity of monocular RGB images, multi-camera systems leverage synchronized, overlapping viewpoints. These approaches fall into two families. BEV-based methods transform perspective-view features into a bird's-eye-view grid via discretized pooling \cite{oft} or learned depth distributions \cite{lss,bevdet}. Query-based methods \cite{detr3d,petr,focalpetr,sparse4d,sparse4dv2,sparse4dv3} instead let learnable object queries attend directly to multi-view features, often using 3D positional encodings \cite{petr,focalpetr,streampetr}. Temporal pipelines further stabilize predictions through BEV warping \cite{bevdet4d,bevformer}, spatiotemporal query attention \cite{petrv2,sparse4d,sparse4dv2,sparse4dv3}, or memory banks \cite{streampetr}. While effective, these approaches rely on calibrated multi-camera rigs; we instead focus on the more easily deployable single-camera setting.

\subsubsection{Monocular 3D Detection.} Monocular systems are cheap and straightforward to deploy, but must recover 3D structure from a single image without additional geometric cues (\eg, LiDAR or multi-view). To handle this ill-posed depth problem, one category of methods leverages LiDAR measurements for auxiliary training supervision to guide representation learning \cite{monotakd,occupancym3d}. Another family of frameworks enforces explicit geometric priors relating object size, depth, and the ground plane \cite{mogde,monocd,monodgp}. However, these handcrafted assumptions are often restricted to flat ground scenarios and specific environments. Other architectures rely solely on direct 3D bounding box supervision \cite{meier2025lead,gate3d,smoke,objectaspoints,fcos3d}. Our framework follows this unconstrained setting, bypassing restrictive assumptions to maximize flexibility while exploiting monocular temporal cues to enhance detection accuracy. Existing temporal approaches \cite{kinematic,monotdf,quasimono}, however, assume a fixed deployment domain, leaving their behavior under zero-shot domain shift unexamined. Instead, we explicitly aim for strong domain generalization.

\subsubsection{Domain Generalization in 3D Perception.} Domain generalization (DG) targets stability under unseen domain shifts such as different cameras and visual conditions. In 2D tasks, this is often addressed via augmentations or invariant representation learning \cite{domainbed,clipthegap}. Similarly, LiDAR-based 3D detectors counter sensor-induced domain shifts by learning sparsity-invariant features \cite{lsf} or applying adversarial augmentations \cite{3dvfield,lidar_generalization_study}. For multi-view camera systems, approaches such as DG-BEV \cite{dgbev} counteract camera-induced shifts by decoupling depth prediction from pixel size and applying perspective augmentations. Other methods ensure robustness by enforcing consistency across either spatially overlapping cameras \cite{unified_mv_dg} or temporal sequences \cite{tsmabev}. However, as these approaches require multi-view data, they cannot be applied directly to our monocular setup.

In the monocular setting, many approaches rely on unsupervised domain or test-time adaptation, requiring target data during training \cite{stmono3d,dgmono3d,monoct} or inference \cite{monotta}. In contrast, we focus on strict zero-shot domain generalization without access to target data. To the best of our knowledge, MonoGDG \cite{MonoGDG} is the only monocular domain generalization method. While accounting for variations in appearance and camera configurations, MonoGDG is restricted to single frames and does not use multi-frame cues. We instead use monocular multi-frame cues.

\section{A Cross-Domain Benchmark for Monocular Temporal 3D Detection} \label{sec:datasets}

\begin{table}[t]
\centering
\caption{\textbf{Cross-dataset benchmark overview.} Our cross-dataset domain generalization benchmark comprises four 3D monocular detection datasets, differing markedly in their field-of-view (FOV), intrinsics, resolution \& geographic coverage.}
\label{tab:datasets}
\renewcommand{\arraystretch}{1.1}
\setlength{\tabcolsep}{3.5pt}
\scriptsize
\begin{tabularx}{\linewidth}{@{}Xcccc}
\toprule
\textbf{Property} & \textbf{nuScenes}\cite{nuscene} & \textbf{Waymo}\cite{waymo} & \textbf{Lyft L5}\cite{lyft} & \textbf{ONCE}\cite{once} \\
\midrule
Annotated frames (train / val)
& \num{28130} / \num{6019}
& \num{53000} / \num{13280}
& \num{15498} / \num{3150}
& \num{5000} / \num{3000} \\

Front camera resolution (in px)
& 1600\,$\times$\,900
& 1920\,$\times$\,1280
& 1224\,$\times$\,1024
& 1920\,$\times$\,1020 \\

Front focal length $f_x$ (in px)
& 1266
& 2059
& 844
& 960 \\

Front horizontal FOV (in °)
& 70.0
& 50.0
& 70.0
& 90.0 \\

Number of object categories
& 10
& 4
& 9
& 5 \\

Country
& USA \& Singapore
& USA
& USA
& China \\

\bottomrule
\end{tabularx}
\end{table}
\begin{figure}[t]
    \centering
    \pgfplotsset{
    colormap={waymo}{
         RGB(0pt)=(255,245,235);
         RGB(1pt)=(253,167,98);
         RGB(2pt)=(127,39,4);
    }
}

\pgfplotsset{
    colormap={nuscenes}{
         RGB(0pt)=(247,251,255);
         RGB(1pt)=(176,210,232);
         RGB(2pt)=(8,48,107);
    }
}

\pgfplotsset{
    colormap={once}{
         RGB(0pt)=(252,251,253);
         RGB(1pt)=(146,142,195);
         RGB(2pt)=(74,20,134);
    }
}

\pgfplotsset{
    colormap={lyft}{
         RGB(0pt)=(247,252,245);
         RGB(1pt)=(93,185,107);
         RGB(2pt)=(0,68,27);
    }
}

\!\!\begin{tikzpicture}[clip, every node/.style={font=\sffamily\small}]

\begin{groupplot}[
    group style={
        group size=4 by 1,
        horizontal sep=0.5cm,
    },
    title style={
        yshift=-7.0pt,
    },
    height=4.1cm,
    xtick pos=bottom,
    ytick pos=left,
    xlabel style={
        yshift=2pt,
    },
    ylabel style={
        yshift=-4pt,
    },
    xmin=-45, xmax=45,
    ymin=0, ymax=100,
    enlargelimits=false,
    axis equal image,
    axis on top,
    xtick={
            -40, -20, 0, 20, 40
        },
    xticklabels={
        -40, -20, 0, 20, 40
    },
    colormap name=waymo,
    colorbar,
    colorbar style={
        ytick={1, 9},
        yticklabels={L, \textcolor{white}{H}},
        yticklabel style={
            xshift=-9.95pt,
        },
        major tick length=0pt,
        xtick=\empty,
        xshift=-9.1pt,
    },
    point meta min=0,
    point meta max=10,
]

\nextgroupplot[title={Waymo}, ylabel={y (in m)}, ytick={
            0, 20, 40, 60, 80, 100
        },
    yticklabels={
        0, 20, 40, 60, 80, 100
    }, colormap name=waymo]
\addplot graphics [
    xmin=-45,
    xmax=45,
    ymin=0,
    ymax=100,
] {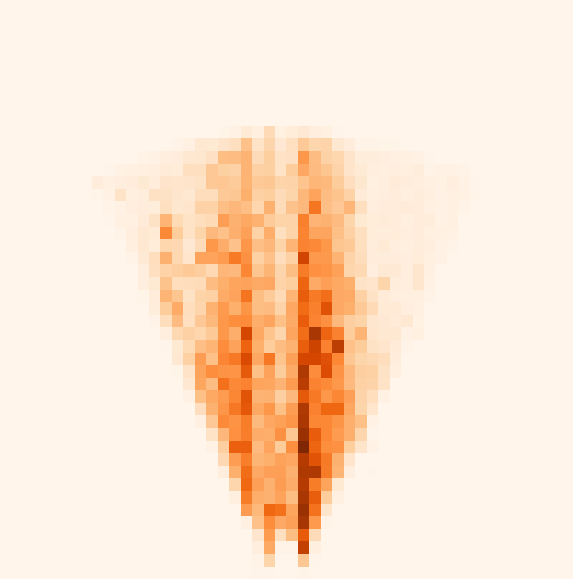};

\nextgroupplot[title={nuScenes\vphantom{y}}, ytick={}, yticklabels={}, colormap name=nuscenes,]
\addplot graphics [
includegraphics cmd=\pgfimage,
    xmin=-45,
    xmax=45,
    ymin=0,
    ymax=100,
] {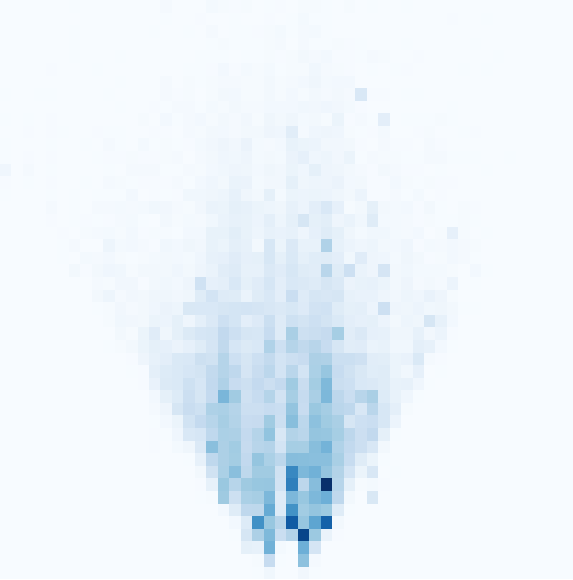};

\nextgroupplot[title={Lyft}, ytick={}, yticklabels={}, colormap name=lyft,]
\addplot graphics [
    xmin=-45,
    xmax=45,
    ymin=0,
    ymax=100,
] {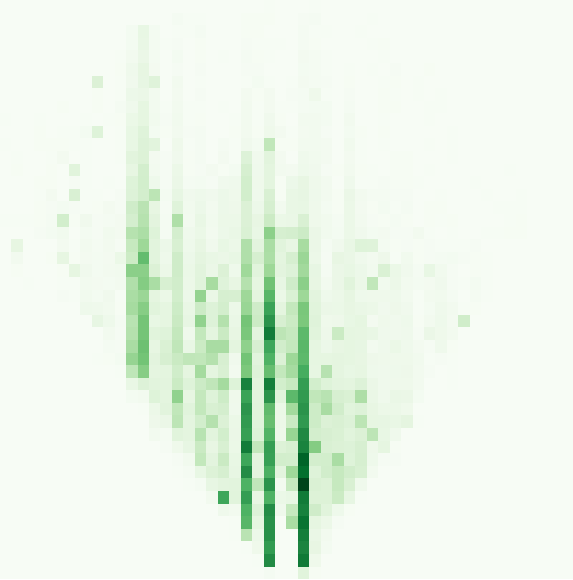};

\nextgroupplot[title={ONCE\vphantom{y}}, ytick={}, yticklabels={},colormap name=once,]
\addplot graphics [
    xmin=-45,
    xmax=45,
    ymin=0,
    ymax=100,
] {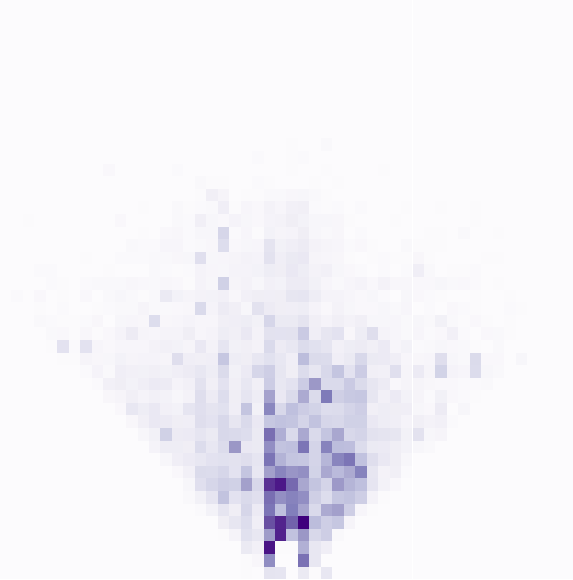};

\end{groupplot}

\node[anchor=center] at (5.3, -0.635) {x (in m)};

\end{tikzpicture}
    \caption{\textbf{Object locations across datasets.} Top-view heatmaps of annotated object locations for Waymo~\cite{waymo}, nuScenes~\cite{nuscene}, Lyft L5~\cite{lyft}, and ONCE~\cite{once}. A larger saturation corresponds to a higher density. Object locations differ significantly between datasets.}
    \label{fig:object_dist}
\end{figure}

Existing monocular domain generalization benchmarks focus strictly on single-frame 3D detection \cite{MonoGDG,stmono3d,monoct}. To evaluate monocular temporal 3D detection under domain shifts, we present a new cross-dataset benchmark (\cf \cref{tab:datasets}), adapting four autonomous driving datasets. In particular, we employ nuScenes \cite{nuscene}, Waymo \cite{waymo}, Lyft \cite{lyft}, and ONCE \cite{once}. %

\subsubsection{Training and Validation Protocol.} We train separate models independently on the training splits of nuScenes \cite{nuscene}, Waymo \cite{waymo}, and Lyft \cite{lyft}. We evaluate each model on the validation splits of all four datasets. Note, ONCE \cite{once} serves only as an unseen validation dataset. This allows us to measure both in-domain accuracy and, more importantly, cross-domain generalization over three different datasets. As depicted in \cref{fig:object_dist}, the spatial distribution of objects varies significantly across the different datasets, providing a significant challenge for generalization.

\subsubsection{Additional Details.} As Waymo is recorded using a high frame rate, we follow existing work \cite{meier2025lead,cadnn} and subsample frames. In particular, we sample every third frame. As all datasets use different object-category taxonomies (\cf \cref{tab:datasets}), we adopt a shared taxonomy comprising ``vehicle'', ``pedestrian'', and ``bicycle''. Note that we focus on generalizable 3D detections; considering an open-vocabulary setting for object categories is left to future work.

\subsubsection{Evaluation Metrics.} We employ the standard nuScenes protocol \cite{nuscene}, reporting mean Average Precision (mAP) alongside the composite nuScenes Detection Score (NDS). NDS serves as our primary metric, combining ground-plane distance-based mAP with true-positive error attributes covering translation, scale, and orientation. To measure overall cross-domain generalization, we average mAP and NDS over all datasets, excluding the training domain.

\section{Method: MAGneT-3D}
\label{sec:method}

\subsubsection{Problem Definition.} We approach \emph{monocular} and \emph{temporal monocular} 3D object detection under \emph{single-domain generalization}. At each timestep $t\in\{1,\ldots, T\}$, given an RGB image $\mathbf{x}_t \in \mathbb{R}^{\rm H \times \rm W \times 3}$ and camera intrinsics $\mathbf{K}_t\in\mathbb{R}^{3\times 3}$, the goal is to predict a set of oriented 3D bounding boxes $\mathbf{y}_t = \{\mathbf{b}_m\}_{m=1}^{M}$, each parameterized as $\mathbf{b} = (k, \mathbf{c}, \mathbf{d}, \theta)$, where $k$ is the class label, $\mathbf{c} = (c_x, c_y, c_z)^\top$ the 3D center in camera coordinates, $\mathbf{d} = (d_x, d_y, d_z)^\top$ the box dimensions, and $\theta$ the yaw angle.
In the temporal setting, inputs arrive as a sequence of $T$ frames $\mathcal{S} = \{(\mathbf{x}_t, \mathbf{K}_t, \mathbf{E}_t)\}_{t=1}^{T}$, where $\mathbf{E}_t \in \operatorname{SE}(3)$ is the camera ego-pose at timestep $t$. A causal detector produces per-frame predictions $\hat{\mathbf{y}}_t$ from past and present observations. 
Under \emph{single-domain} generalization, we train exclusively on a single labeled source domain $\mathcal{D}_S = \{(\mathbf{x}_i, \mathbf{K}_i, \mathbf{y}_i)\}_{i=1}^{N_S}$ and aim to learn a monocular temporal 3D detector that generalizes to a completely \emph{unseen} target domain $\mathcal{D}_T = \{(\mathbf{x}_j, \mathbf{K}_j)\}_{j=1}^{N_T}$ at test-time.

\begin{figure}[t]
    \centering
    \input{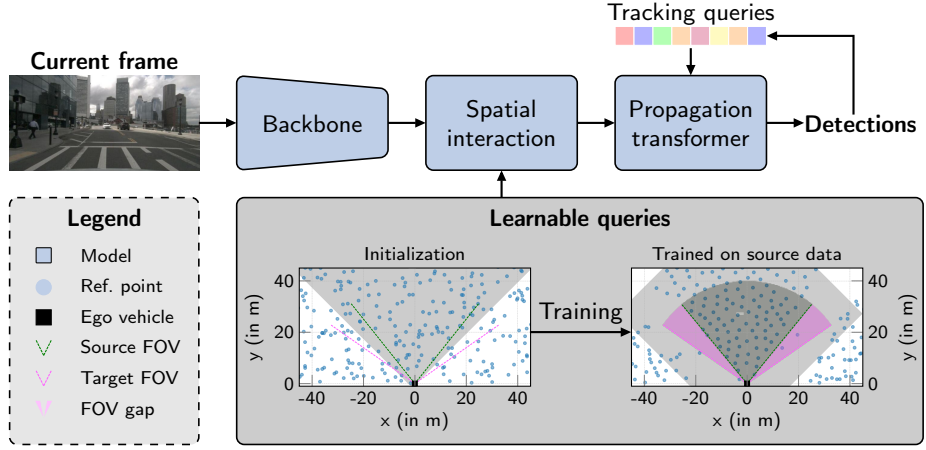}
    \caption{\textbf{Query-based 3D detectors \cite{streampetr,sparse4d,sparse4dv2,sparse4dv3}.} The current image is fed into a backbone. The resulting image features interact with the learnable queries \emph{(bottom)}. A propagation transformer takes in query features and historically tracked queries, producing refined 3D detections. Queries \emph{(bottom)} are learned during training but static during inference and introduce training dataset dependent biases, introducing issues under domain shifts, \eg, when the field-of-view (FOV) is changing.}
    \label{fig:query_based}
\end{figure}

\subsection{Preliminaries: Limitations of Query-Based Detection}
Our method builds upon query-based 3D object detectors \cite{streampetr,sparse4d,sparse4dv3,sparse4dv2,petr}. These represent potential objects as a fixed set of 3D queries. These queries are learnable during training but fixed during inference. Each query encodes a candidate position in 3D space (\ie, a bird's-eye-view reference point) together with a feature vector. As shown in \cref{fig:query_based}, image features are extracted by a 2D backbone. Since queries reside in 3D and image features in 2D, a spatial interaction module brings them into a common space: either by enriching image features with 3D positional encodings so queries attend to them directly \cite{petr,streampetr}, or by projecting each query's reference point into the image plane and sampling features there \cite{sparse4d,sparse4dv3}. A transformer decoder then uses the resulting query-feature correspondences to refine each query's 3D position, dimensions, and class score. For temporal modeling, past query states are stored in a memory bank and fused with the current queries. While these models achieve strong in-domain performance, their domain generalization is limited, as we discuss next.

\subsubsection{Limitations of Query-Based Detectors.}
\begin{figure}[t]
    \centering
    \input{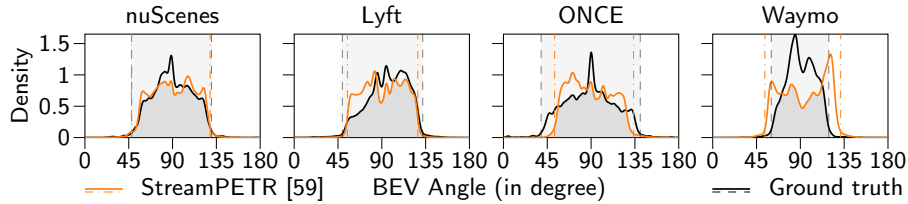}
    \vspace{-14pt}
    \caption{\textbf{Learnable, static queries bias analysis.} We visualize the angle distributions of ground truth object detections \emph{vs.} StreamPETR~\cite{streampetr} predictions trained on nuScenes~\cite{nuscene}. Dashed lines indicate the 1\textsuperscript{st} and 99\textsuperscript{th} percentiles of both distributions. StreamPETR predictions are biased to the nuScenes FOV across all target datasets.}
    \label{fig:streampetr_one_d_fov}
\end{figure}

Prior work attributes cross-dataset drops to domain shifts in camera configurations, annotations, or image appearance \cite{dgmono3d,MonoGDG,monoct}. Beyond these, we identify a failure mode rooted in the query initialization itself, caused by underlying spatial distribution shifts across datasets. Training biases query reference points toward source-domain ground-truth locations (\cf \cref{fig:object_dist}). We observe this phenomenon, for example, in the field-of-view (FOV), where queries migrate away from peripheral regions and cluster within the source viewing boundaries (\cf \cref{fig:query_based}). Although the decoder refinement head can adjust these positions, it typically performs only minor, incremental refinements. 
Because the underlying initialization queries are biased and remain static during inference, they cannot adapt to cover regions outside the source-domain range. Consequently, predictions in the target domain remain largely constrained to the source angular distribution despite the domain shift (\cf \cref{fig:streampetr_one_d_fov}). This limitation significantly hinders domain generalization by preventing the model from adapting its spatial representation to unseen target-domain configurations.

\subsection{Overview of \methodsymbol{}}
\begin{figure}[t]
    \centering
    \input{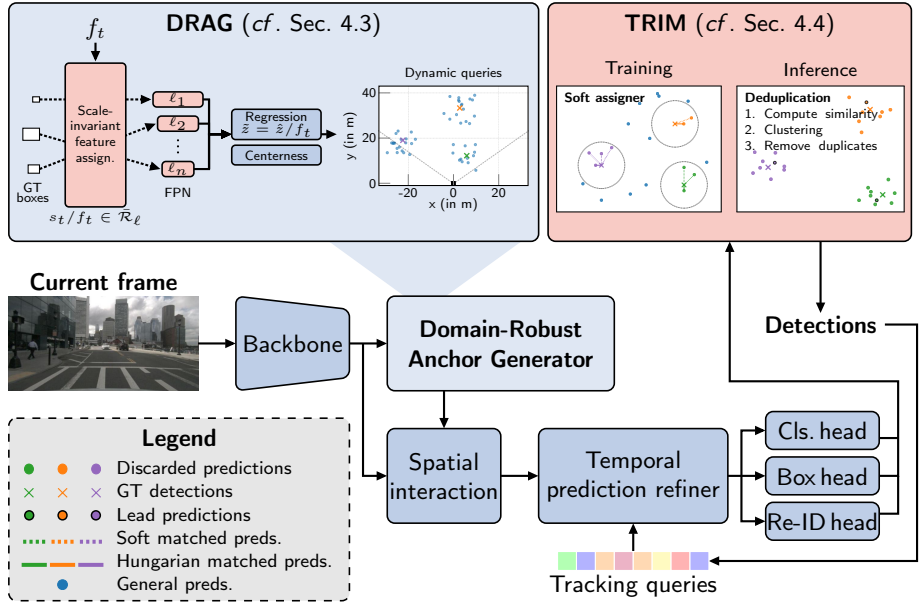}
    \caption{\textbf{\methodsymbol{}.} Instead of static, learnable queries, we generate dynamic proposals \emph{(left top)} based on image features (\anchorgensymbol{}; \cf \cref{sec:cag}), eliminating source domain biases. The proposals interact spatially with the image features and are fed, alongside historic queries, into the temporal prediction refiner. During training, our soft assigner matches each target to multiple proposals for dense supervision. During inference, duplicate objects are merged via clustering (TRIM; \cf \cref{sec:tpr}).}
    \label{fig:main_architecture}
\end{figure}

To overcome the limitations associated with static, learned queries, we present \methodsymbol{}. By introducing a dynamic initialization routine, \methodsymbol{} moves beyond static queries, avoiding target dataset biases and providing a strong domain generalization. On a high level, \methodsymbol{} builds on StreamPETR \cite{streampetr} and consists of two novel components: \emph{(i)} Our Domain-Robust Anchor Generator (\anchorgensymbol) extracts image features to construct per-frame 3D proposals that naturally adapt to different camera setups, and \emph{(ii)} our Temporal Refinement and Identity Merging (\propgationtransformersymbol) refines these proposals via soft assignment and dense supervision during training, paired with contrastive clustering-based deduplication at inference. An overview of \methodsymbol{} is in \cref{fig:main_architecture}.

\subsection{\anchorgen{} (\anchorgensymbol{})}
\label{sec:cag}
We introduce \textbf{D}omain-\textbf{R}obust \textbf{A}nchor \textbf{G}enerator (\anchorgensymbol{}). Instead of static, learned queries that are prone to biases, \anchorgensymbol{} dynamically generates per-frame 3D anchors that adapt to any target camera configuration (\cf \cref{fig:main_architecture}). \anchorgensymbol{} achieves this by extracting high-confidence regions directly from the feature map to derive initial 3D position estimates. Implemented as a lightweight convolutional network, \anchorgensymbol{} provides robust generalization across varying image resolutions and is trained using intermediate supervision. Crucially, this setup allows us to decompose the 3D position priors into projected 2D centers and a corresponding depth estimate, which are then backprojected to form the final 3D anchors. Specifically, we append an FCOS3D head \cite{fcos3d} as a plug-and-play module directly onto our backbone and use its standard losses and heads.

In particular, given a predicted projected center $\hat{\mathbf{p}}_t = (\hat{u}_t, \hat{v}_t, 1)^\top$ and its predicted depth $\hat{z}_t$ at timestep $t$, we recover the initial 3D center estimate $\hat{\mathbf{c}}_t$ via backprojection using the current camera intrinsics $\mathbf{K}_t$ by
$\hat{\mathbf{c}}_t = [\hat{x}_t, \hat{y}_t, \hat{z}_t]^\top = \hat{z}_t\,\mathbf{K}_t^{-1}\hat{\mathbf{p}}_t
$. %
The estimated center $\hat{\mathbf{c}}_t$ serves as the input spatial prior for the subsequent temporal prediction refiner. To maximize the out-of-domain robustness, we introduce two structural adjustments targeting scale and intrinsic variations.

\subsubsection{Virtual Depth Estimation}
While subsequent layers of the temporal prediction refiner directly receive the camera intrinsics $\mathbf{K}_t$ to handle geometric shifts, the backbone does not. Consequently, forcing the network to regress metric depth directly creates a significant sensitivity to focal-length shifts. To debias the visual representation from these intrinsic variations, we predict a normalized virtual depth \cite{omni3d} $\tilde{z}_t = \hat{z}_t\,/\,f_t$ instead of absolute metric depth. Here, $f_t\in\mathbb{R}$ represents the current focal length. Metric depth required for 3D box construction is recovered at inference-time via $\hat{z}_t = f_t \tilde{z}_t$.

\subsubsection{Scale-Invariant Feature Assignment.}
A standard feature pyramid network (FPN) produces a pyramid of $L$ feature maps at multiple resolutions. During training, objects are assigned to a specific level $\ell\in\{1, \ldots, L\}$ (\ie, resolution stage) of the feature pyramid dependent on the 2D object size. However, 2D object size depends both on its 3D size and the camera configuration (foremost the focal length). To improve domain generalization and ensure cross-camera consistency, we seek to make this assignment independent of the focal length.

In particular, instead of using 2D object size for level-assignment, we propose to perform level-assignment in a focal-normalized space. We divide the standard pixel boundaries for level-assignment by a fixed reference focal length $f_{\mathrm{ref}}$ (set to the nuScenes focal length, \SI{1266}{px}), yielding the normalized interval $\bar{\mathcal{R}}_\ell = [r_\ell^{\min}/f_{\mathrm{ref}},\, r_\ell^{\max}/f_{\mathrm{ref}}]$. For a proposal observed under mean focal length $f_t = (f_{x,t} + f_{y,t})\,/\,2$, we rescale this interval back to pixel units by $f_t$ and assign the proposal to level $\ell$ only if its pixel scale $s_t$ satisfies $s_t \in f_t\,\bar{\mathcal{R}}_\ell$. Equivalently, the focal-normalized scale $s_t / f_t$ must fall within $\bar{\mathcal{R}}_\ell$. The center-sampling radius $\rho_\ell$ is rescaled by $f_t / f_{\mathrm{ref}}$ analogously. Because $s_t / f_t$ is independent of the camera focal length, this guarantees that an instance always routes to the same feature layer regardless of camera intrinsics and other resizing transforms. This enables the network to obtain a holistic understanding of the scene rather than local pixel sizes, significantly boosting domain generalization.

\subsection{\propgationtransformer{} (\propgationtransformersymbol{})}
\label{sec:tpr}
To map dense proposals to temporal object instances across environments, we introduce \textbf{T}emporal \textbf{R}efinement and \textbf{I}dentity \textbf{M}erging (\propgationtransformersymbol{}). The dynamically generated 3D anchors from \anchorgensymbol{} serve as explicit spatial priors, shifting the role of the refiner from coarse localization to fine residual refinement. To fully exploit these high-quality proposals, we replace one-to-one assignments with a dense supervision strategy during training (\cf \cref{fig:main_architecture}), paired with an inference-time deduplication stage to filter redundant predictions.

\subsubsection{Soft Assignment for Dense Supervision.}
\label{sec:soft_assignment}
During training, \anchorgensymbol{} can still entail biases from the training domain, \ie, producing proposals that lie exceptionally close to ground-truth objects. Consequently, standard one-to-one Hungarian matching used by StreamPETR \cite{streampetr} maps each ground-truth object to only the single most confident anchor. Because this anchor is already accurate, the decoder refinement head receives immediate confirmation from the loss to perform only minor adjustments. This creates a severe mismatch between training and inference. During inference on an unseen target domain, the generated anchors are inherently of lower quality due to distribution shifts, yet the refinement head was trained exclusively on anchors that require minimal correction.

To encourage the decoder to learn robust, large corrections, we introduce a soft assignment strategy during training. We map up to $M_{\text{assign}}$ queries to each ground-truth object based on matching quality, while enforcing a radial constraint $r_{\max}$ to exclude distant proposals. Supervising a wider distribution of nearby queries provides a more diverse learning signal, teaching the decoder to generalize its refinement capabilities beyond near-perfect inputs. \Cref{fig:main_architecture} \emph{(top right)} contrasts this with standard one-to-one assignment.

\subsubsection{Clustering-Based Deduplication.}
\label{sec:clustering}
While dense anchor generation paired with soft assignment improves robustness, it yields multiple high-confidence duplicate predictions per object at inference, which significantly hampers precision. We address this by coupling a contrastive learning objective during training with a clustering pipeline at test time.

During training, we project each prediction into an object identity space and supervise the resulting object identity vectors (\ie, embeddings in latent space) using a supervised contrastive loss $\mathcal{L}_{\text{\propgationtransformersymbol}}$. Let $I_t = \{1, \dots, M_t\}$ be the set of indices for all predictions in $\hat{\mathbf{y}}_t$ at timestep $t$. We define $A_t(i) \equiv \{j \in I_t \setminus \{i\} | o_j = o_i\}$ as the index set of all other predictions mapping to the same ground-truth object identity $o_i$ as prediction $i$. Following SupCon~\cite{supcon}, the clustering loss is:
\begin{equation}
\mathcal{L}_{\text{\propgationtransformersymbol}} = -\sum_{i \in I_t} \frac{1}{|A_t(i)|} \sum_{j \in A_t(i)} \log \frac{\exp(\mathbf{v}_i \cdot \mathbf{v}_j / \tau)}{\sum_{m \in I_t \setminus \{i\}} \exp(\mathbf{v}_i \cdot \mathbf{v}_m / \tau)},
\label{eq:trim}
\end{equation}
where $\mathbf{v_{\cdot}}$ denotes the normalized identity vector. $\tau>0$ is the temperature parameter. This objective pulls embeddings of the same object together while pushing different objects apart. At test time, predictions with similar identity vectors are grouped into spatial clusters using these learned embeddings. To eliminate redundant detections without sacrificing recall, we retain only the single (\ie, $N_{\mathrm{keep}}=1$) highest-confidence prediction per cluster (\cf \cref{fig:main_architecture} \emph{(top right)}).

\subsubsection{Discussion.} Both \anchorgensymbol{} and \propgationtransformersymbol{} act in synergy. \anchorgensymbol{} substitutes overfit static queries with dense, scene-adaptive anchors. The soft assignment in \propgationtransformersymbol{} leverages this density to construct a more demanding, robust training signal, and clustering-based deduplication eliminates redundant predictions at inference. Together, they enable robust detection across severe distribution shifts without compromising precision.

\subsubsection{Full Training Objective.}
We adopt a two-stage training strategy to maximize robustness. We first warm up the network by training the anchor generator via $\mathcal{L}_{\mathrm{FCOS3D}}$, and then optimize the entire framework end-to-end. This provides stable 3D spatial priors to the refinement head from the start of joint training. The total optimization objective $\mathcal{L}$ sums the individual multitask components:
$
\mathcal{L} = \mathcal{L}_{\mathrm{FCOS3D}} + \mathcal{L}_{\mathrm{StreamPETR}} + \mathcal{L}_{\mathrm{TRIM}},
$
where $\mathcal{L}_{\mathrm{FCOS3D}}$ is the standard FCOS3D loss \cite{fcos3d} computed on our virtual depth and using scale-invariant assignment. $\mathcal{L}_{\mathrm{StreamPETR}}$ is the standard StreamPETR loss \cite{streampetr} under our soft assignment, and $\mathcal{L}_{\mathrm{TRIM}}$ is the supervised contrastive loss (\cf \cref{eq:trim}). %

\definecolor{bestcolor}{RGB}{198,239,206}    %
\definecolor{secondcolor}{RGB}{255,235,156} %
\begin{table*}[t]
  \centering
  \fontsize{4pt}{5pt}\selectfont
  \setlength{\tabcolsep}{1.125pt} %
    \renewcommand{\arraystretch}{1.1}
  \caption{\textbf{Comparison against the state of the art.} Each model is trained on a single source dataset \emph{(vertical)} and evaluated on all target datasets \emph{(horizontal)}. Avg.\,Cross is the mean over the three unseen datasets. The best result per training dataset is highlighted in green \colorindicator{bestcolor}; the second best in yellow \colorindicator{secondcolor}; in-domain results in gray \colorindicator{gray}. mAP \& NDS are both in \% ($\uparrow$). nuScenes is abbreviated by nuSCN.}
  \label{tab:combined_results}
  \begin{tabularx}{\linewidth}{ll S[table-format=2.1]S[table-format=2.1]S[table-format=2.1]S[table-format=2.1]S[table-format=2.1] S[table-format=2.1]S[table-format=2.1]S[table-format=2.1]S[table-format=2.1]S[table-format=2.1]}
    \toprule
    \raisebox{-2.2pt}[0pt][0pt]{\multirow{2}{*}{\textbf{Method}}} & \raisebox{-1.65pt}[0pt][0pt]{\textbf{Val\,$\xrightarrow{}$}} & \multicolumn{5}{c}{\textbf{mAP\,$\uparrow$}} & \multicolumn{5}{c}{\textbf{NDS\,$\uparrow$}} \\
    \cmidrule(lr){3-7} \cmidrule(lr){8-12}
    & \raisebox{1.65pt}[0pt][0pt]{\textbf{Train}\,$\downarrow$} & \textbf{nuSCN} & \textbf{Waymo} & \textbf{Lyft} & \textbf{ONCE} & \textbf{Avg.\,Cross\,} & \textbf{nuSCN} & \textbf{Waymo} & \textbf{Lyft} & \textbf{ONCE} & \textbf{Avg.\,Cross} \\
    \midrule

    \multirow{3}{*}{StreamPETR \cite{streampetr}} & nuSCN & \color{gray} 26.5 & 4.8 & 5.0 & 8.2 & 6.0 & \color{gray} 28.5 & \cellcolor{secondcolor} 13.1 & 10.6 & 12.7 & 12.1 \\
    & Waymo & 4.2 & \color{gray} \cellcolor{secondcolor} 34.5 & 0.8 & \cellcolor{secondcolor} 5.4 & 3.4 & \cellcolor{secondcolor} 9.3 & \color{gray} 38.0 & \cellcolor{secondcolor} 8.9 & 10.2 & \cellcolor{secondcolor} 9.5 \\
    & Lyft & 1.0 & 0.0 & \color{gray} 27.8 & 5.4 & 2.1 & 5.4 & 4.4 & \color{gray} 32.6 & 11.2 & 7.0 \\
    \midrule

    \multirow{3}{*}{Sparse4D v3 \cite{sparse4dv3}} & nuSCN & \color{gray} 24.8 & 4.4 & \cellcolor{secondcolor} 8.6 & \cellcolor{secondcolor} 11.1 & \cellcolor{secondcolor} 8.0 & \color{gray} 27.0 & 9.5 & \cellcolor{secondcolor} 13.1 & \cellcolor{secondcolor} 14.7 & \cellcolor{secondcolor} 12.4 \\
    & Waymo & \cellcolor{secondcolor} 4.9 & \color{gray} 32.8 & \cellcolor{secondcolor} 2.0 & 4.8 & \cellcolor{secondcolor} 3.9 & 8.9 & \color{gray} 38.9 & 7.8 & 9.1 & 8.6 \\
    & Lyft & \cellcolor{secondcolor} 2.2 & 1.5 & \color{gray} \cellcolor{secondcolor} 32.0 & 3.8 & 2.5 & 6.2 & 6.1 & \color{gray} \cellcolor{bestcolor} 37.9 & 7.4 & 6.6 \\
    \midrule

    \multirow{3}{*}{Far3D \cite{far3d}} & nuSCN & \color{gray} \cellcolor{secondcolor} 28.3 & 2.3 & 5.6 & 9.3 & 5.7 & \color{gray} \cellcolor{secondcolor} 29.8 & 10.8 & 12.8 & 13.2 & 12.3 \\
    & Waymo & 1.9 & \color{gray} \cellcolor{bestcolor} 36.9 & 0.3 & 2.6 & 1.6 & 7.7 & \color{gray} \cellcolor{bestcolor} 39.6 & 7.1 & 7.7 & 7.5 \\
    & Lyft & 1.3 & 1.1 & \color{gray} 31.0 & \cellcolor{bestcolor} 6.8 & 3.0 & \cellcolor{secondcolor} 7.3 & 8.4 & \color{gray} 33.0 & 10.7 & 8.8 \\
    \midrule

    \multirow{3}{*}{BEVFormer v2 \cite{bevformerv2}} & nuSCN & \color{gray} 21.8 & \cellcolor{secondcolor} 6.2 & 2.5 & 10.6 & 6.4 & \color{gray} 22.2 & 11.2 & 9.1 & 14.5 & 11.6 \\
    & Waymo & 2.1 & \color{gray} 11.5 & 1.8 & 5.2 & 3.0 & 6.4 & \color{gray} 15.8 & 8.3 & \cellcolor{secondcolor} 10.7 & 8.4 \\
    & Lyft & 1.4 & \cellcolor{secondcolor} 3.8 & \color{gray} 19.4 & 5.5 & \cellcolor{secondcolor} 3.6 & 4.2 & \cellcolor{secondcolor} 10.7 & \color{gray} 26.0 & \cellcolor{secondcolor} 11.8 & \cellcolor{secondcolor} 8.9 \\
    \midrule

    \multirow{3}{*}{\methodsymbol(Ours)} & nuSCN & \color{gray} \cellcolor{bestcolor} 28.5 & \cellcolor{bestcolor} 10.1 & \cellcolor{bestcolor} 14.3 & \cellcolor{bestcolor} 12.4 & \cellcolor{bestcolor} 12.3 & \color{gray} \cellcolor{bestcolor} 32.9 & \cellcolor{bestcolor} 17.2 & \cellcolor{bestcolor} 19.5 & \cellcolor{bestcolor} 19.1 & \cellcolor{bestcolor} 18.6 \\
    & Waymo & \cellcolor{bestcolor} 13.4 & \color{gray} 34.5 & \cellcolor{bestcolor} 13.8 & \cellcolor{bestcolor} 9.2 & \cellcolor{bestcolor} 12.2 & \cellcolor{bestcolor} 18.8 & \color{gray} \cellcolor{secondcolor} 39.3 & \cellcolor{bestcolor} 19.5 & \cellcolor{bestcolor} 14.0 & \cellcolor{bestcolor} 17.4 \\
    & Lyft & \cellcolor{bestcolor} 4.0 & \cellcolor{bestcolor} 5.2 & \color{gray} \cellcolor{bestcolor} 32.5 & \cellcolor{secondcolor} 6.0 & \cellcolor{bestcolor} 5.0 & \cellcolor{bestcolor} 7.7 & \cellcolor{bestcolor} 11.6 & \color{gray} \cellcolor{secondcolor} 34.5 & \cellcolor{bestcolor} 11.9 & \cellcolor{bestcolor} 10.4 \\
    \bottomrule
  \end{tabularx}
\end{table*}

\section{Experiments}
\label{sec:results}

We first compare \methodsymbol{} against state-of-the-art multi-view approaches adapted to monocular data. Next, we ablate the core components of \methodsymbol{}. Finally, we conduct a spatial distribution analysis providing further insights.

\subsubsection{Implementation Details \& Baselines.} All models use a ResNet-50 backbone \cite{resnet} pre-trained on ImageNet \cite{imagenet}. To establish monocular baselines, we employ state-of-the-art multi-view detectors \cite{streampetr,sparse4dv3,bevformerv2} and retrained them in the monocular video setting.
All models are trained using an identical budget of \num{675120} iterations (about \num{24} nuScenes epochs) to ensure a fair comparison across datasets. We use a batch size of \num{16}, a maximum temporal history of \num{20} frames, and AdamW \cite{adamw} with a learning rate of $\eta = 4\times 10^{-4}$ and weight decay of $\lambda = 10^{-2}$. For the soft assignment strategy, we set $M_{\text{assign}}$ to \num{5} and $r_{\max}$ to \SI{5}{m}, while the temperature $\tau$ is set to \num{0.15}.

\subsection{Comparison with State of the Art}
\Cref{tab:combined_results} reports results on our proposed cross-dataset domain generalization benchmark. To cover the major families of camera-based (temporal) 3D detectors, we compare against one state-of-the-art representative from each: StreamPETR \cite{streampetr} (dense query-based with temporal modeling), Sparse4D v3 \cite{sparse4dv3} (sparse query-based), BEVFormer v2 \cite{bevformerv2} (BEV-based), and Far3D \cite{far3d} (long-range query-based). Despite their architectural differences, all four baselines share the same fundamental weakness: strong in-domain accuracy that degrades significantly under domain shifts. Across all training datasets, \methodsymbol achieves the best average cross-dataset NDS and mAP by a significant margin. Specifically, our model improves average cross-dataset NDS by \SI{50}{\%} on nuScenes \cite{nuscene}, \SI{83}{\%} on Waymo \cite{waymo}, and \SI{16}{\%} on Lyft \cite{lyft} relative to the best baseline. While some architectures achieve a higher in-domain accuracy on Waymo (Far3D) and Lyft (Sparse4D v3), they suffer severely outside of their source domain. For example, Far3D trained on Waymo achieves an NDS of only \SI{7.1}{\%} on Lyft. For reference, in the very same setting, \methodsymbol attains \SI{19.5}{\%}.

\subsection{Ablation Study}
\label{sec:ablation}
We conduct ablations to isolate the contribution of \methodsymbol's core components.
Specifically, we ablate/analyze \emph{(i)} the overall model configuration relative to the baseline (\ie, StreamPETR\cite{streampetr}), \emph{(ii)} the design choices within \anchorgensymbol, and \emph{(iii)} \propgationtransformersymbol.
In all ablations, we train on the nuScenes training set \cite{nuscene} and report detection accuracy on all four validation sets.

\subsubsection{Main Ablation.}
\Cref{tab:ablation_main} isolates the impact of each \methodsymbol component over the baseline (config.\ 1). Integrating \anchorgensymbol alone (config.\ 2) improves the average cross-dataset accuracy from \num{12.1} to \SI{14.7}{\%} in NDS, but reduces source performance from \SI{28.5}{\%} to \SI{25.6}{\%} in NDS. This drop occurs because matching ground-truth objects to a single anchor limits target diversity, which dilutes the training signal needed to refine boxes. Conversely, deploying \propgationtransformersymbol alone (config.\ 3) leverages multi-hypothesis supervision and increases the average cross-dataset NDS to \SI{13.9}{\%}, though slightly reducing source accuracy on nuScenes to \SI{27.0}{\%}. Combining both components in our full \methodsymbol configuration (config.\ 4) yields the optimal trade-off, maximizing out-of-domain robustness to an average cross-dataset NDS of \SI{18.6}{\%}, while simultaneously increasing source accuracy to \SI{32.9}{\%}. This demonstrates that the synergy between both \anchorgensymbol and \propgationtransformersymbol is required for strong generalization.

\begin{table*}[t]
\centering
\caption{\textbf{Main ablation.} We start with the StreamPETR baseline ablate both \anchorgensymbol{} and \propgationtransformersymbol{}. We train on nuScenes~\cite{nuscene} (nuSCN). mAP \& NDS are both in \% ($\uparrow$). The best result is highlighted in \colorindicator{bestcolor}; the second best in \colorindicator{secondcolor}; in-domain results in \colorindicator{gray}.}
\label{tab:ablation_main}
\fontsize{4pt}{5pt}\selectfont
\setlength{\tabcolsep}{0.75pt} %
\renewcommand{\arraystretch}{1.1}
\begin{tabularx}{\linewidth}{lcc S[table-format=2.1]S[table-format=2.1]S[table-format=2.1]S[table-format=2.1]S[table-format=2.1] S[table-format=2.1]S[table-format=2.1]S[table-format=2.1]S[table-format=2.1]S[table-format=2.1]} %
\toprule
\raisebox{-2.2pt}[0pt][0pt]{\multirow{2}{*}{\textbf{Configuration}}} &
\raisebox{-2.2pt}[0pt][0pt]{\multirow{2}{*}{\textbf{\anchorgensymbol{}}}} &
\raisebox{-2.2pt}[0pt][0pt]{\multirow{2}{*}{\textbf{\propgationtransformersymbol}}}
& \multicolumn{5}{c}{\textbf{mAP\,$\uparrow$}}
& \multicolumn{5}{c}{\textbf{NDS\,$\uparrow$}} \\
\cmidrule(lr){4-8} \cmidrule(lr){9-13} %
& & & \textbf{nuSCN} & \textbf{Waymo} & \textbf{Lyft} & \textbf{ONCE} & \textbf{Avg.\,Cross\,} & \textbf{nuSCN} & \textbf{Waymo} & \textbf{Lyft} & \textbf{ONCE} & \textbf{Avg.\,Cross} \\
\midrule
1.\ StreamPETR & & & \color{gray} 26.5 & 4.8 & 5.0 & 8.2 & 6.0 & \color{gray} \cellcolor{secondcolor}28.5 & 13.1 & 10.6 & 12.7 & 12.1 \\
2. & \cmark & & \color{gray} 20.0 & \cellcolor{secondcolor}7.9 & \cellcolor{secondcolor}8.6 & 8.8 & 8.4 & \color{gray} 25.6 & \cellcolor{secondcolor}16.3 & \cellcolor{secondcolor}13.9 & 13.9 & \cellcolor{secondcolor}14.7 \\
3. & & \cmark & \color{gray} \cellcolor{secondcolor}27.1 & 6.9 & 8.6 & \cellcolor{secondcolor}11.9 & \cellcolor{secondcolor}9.1 & \color{gray} 27.0 & 11.7 & 13.6 & \cellcolor{secondcolor}16.4 & 13.9 \\
4.\ \methodsymbol & \cmark & \cmark & \color{gray} \cellcolor{bestcolor}28.5 &\cellcolor{bestcolor}10.1 &\cellcolor{bestcolor}14.3 & \cellcolor{bestcolor}12.4 & \cellcolor{bestcolor}12.3 & \color{gray} \cellcolor{bestcolor}32.9 & \cellcolor{bestcolor}17.2 & \cellcolor{bestcolor}19.5 & \cellcolor{bestcolor}19.1 & \cellcolor{bestcolor}18.6 \\
\bottomrule
\end{tabularx}%
\end{table*}

\begin{table*}[t]
\centering
\caption{\textbf{\anchorgensymbol{} ablation.} We ablate the core components of \anchorgensymbol{}. Each row removes one component from the full model, trained on nuScenes~\cite{nuscene} (nuSCN). mAP \& NDS are both in \% ($\uparrow$). The best result is highlighted in \colorindicator{bestcolor}; the second best in \colorindicator{secondcolor}; in-domain results in \colorindicator{gray}.}
\label{tab:ablation_generalization}
\fontsize{4pt}{5pt}\selectfont
\setlength{\tabcolsep}{0.78pt}
\renewcommand{\arraystretch}{1.1}
\begin{tabularx}{\linewidth}{lcccccccccc}
\toprule
\raisebox{-2.2pt}[0pt][0pt]{\multirow{2}{*}{\textbf{Setting}}}
& \multicolumn{5}{c}{\textbf{mAP\,$\uparrow$}}
& \multicolumn{5}{c}{\textbf{NDS\,$\uparrow$}} \\
\cmidrule(lr){2-6} \cmidrule(lr){7-11}
& \textbf{nuSCN} & \textbf{Waymo} & \textbf{Lyft} & \textbf{ONCE} & \textbf{Avg.\,Cross\,}
& \textbf{nuSCN} & \textbf{Waymo} & \textbf{Lyft} & \textbf{ONCE} & \textbf{Avg.\,Cross} \\
\midrule
\methodsymbol (Full) & \color{gray} \cellcolor{bestcolor}28.5 &\cellcolor{bestcolor}10.1 & \cellcolor{bestcolor}14.3 &\cellcolor{bestcolor}12.4 &\cellcolor{bestcolor}12.3 & \color{gray} \cellcolor{bestcolor}32.9 & \cellcolor{bestcolor}17.2 &\cellcolor{bestcolor}19.5 & \cellcolor{bestcolor}19.1 & \cellcolor{bestcolor}18.6 \\
\midrule
\;w/o Two-phase training & \color{gray} 26.9 & \cellcolor{secondcolor}8.2 & \cellcolor{secondcolor}12.2 & \cellcolor{secondcolor}12.0 & \cellcolor{secondcolor}10.8 & \color{gray} 30.7 & \cellcolor{secondcolor}16.2 & \cellcolor{secondcolor}18.3 & \cellcolor{secondcolor}18.9 & \cellcolor{secondcolor}17.8 \\
\;w/o Virtual depth & \color{gray} \cellcolor{secondcolor}27.5 & 6.2 & 11.6 & 8.9 & 8.9 & \color{gray} \cellcolor{secondcolor}31.2 & 13.9 & 17.4 & 13.0 & 14.8 \\
\;w/o Scale invariant feat.\ assn. & \color{gray} 17.6 & 6.5 & 5.3 & 7.1 & 6.3 & \color{gray} 23.7 & 14.5 & 12.5 & 11.2 & 12.7 \\
\bottomrule
\end{tabularx}
\end{table*}

\subsubsection{\anchorgensymbol{} Ablation.}
\Cref{tab:ablation_generalization} analyzes the structural choices within \anchorgensymbol{}. Switching to single-stage end-to-end training reduces the average cross-dataset NDS from \SI{18.6}{\%} to \SI{17.8}{\%}. This showcases the benefit of a dedicated anchor warmup training phase. Omitting virtual depth estimation degrades domain generalization from \SI{18.6}{\%} to \SI{14.8}{\%} in NDS. Accuracy on datasets with severe intrinsic shifts, such as Waymo and ONCE, degrades more significantly in this setting. Finally, removing the scale-invariant feature assignment collapses the average cross-dataset score from \SI{18.6}{\%} to just \SI{12.7}{\%} in NDS. Source accuracy also reduces from \SI{32.9}{\%} to \SI{23.7}{\%} in NDS. This demonstrates that scale-invariant assignments are vital to enforce cross-camera consistency.

\begin{table*}[t]
\centering
\caption{\textbf{\propgationtransformersymbol{} analysis.} We analyze the assigner setting and the inference-time clustering of \propgationtransformersymbol{}, trained on nuScenes~\cite{nuscene} (nuSCN). mAP \& NDS are both in \% ($\uparrow$). The best result is highlighted in \colorindicator{bestcolor}; the second best in \colorindicator{secondcolor}; in-domain results in \colorindicator{gray}.}
\label{tab:ablation_clustering}
\setlength{\tabcolsep}{0.4pt}
\fontsize{4pt}{5pt}\selectfont
\renewcommand{\arraystretch}{1.1}
\begin{tabularx}{\linewidth}{lS[table-format=2.1]S[table-format=2.1]S[table-format=2.1]S[table-format=2.1]S[table-format=2.1]S[table-format=2.1]S[table-format=2.1]S[table-format=2.1]S[table-format=2.1]S[table-format=2.1]}
\toprule
\raisebox{-2.2pt}[0pt][0pt]{\multirow{2}{*}{\textbf{Setting}}}
& \multicolumn{5}{c}{\textbf{mAP\,$\uparrow$}}
& \multicolumn{5}{c}{\textbf{NDS\,$\uparrow$}} \\
\cmidrule(lr){2-6} \cmidrule(lr){7-11}
& \textbf{nuSCN} & \textbf{Waymo} & \textbf{Lyft} & \textbf{ONCE} & \textbf{Avg.\,Cross\,}
& \textbf{nuSCN} & \textbf{Waymo} & \textbf{Lyft} & \textbf{ONCE} & \textbf{Avg.\,Cross} \\
\midrule
\methodsymbol (Full, $N_{\mathrm{keep}}\!=\!1$) & \color{gray} \cellcolor{bestcolor}28.5 & \cellcolor{bestcolor}10.1 & \cellcolor{bestcolor}14.3 & \cellcolor{bestcolor}12.4 & \cellcolor{bestcolor}12.3 & \color{gray} \cellcolor{bestcolor}32.9 & \cellcolor{bestcolor}17.2 & \cellcolor{bestcolor}19.5 & \cellcolor{bestcolor}19.1 & \cellcolor{bestcolor}18.6 \\
\midrule
\;w/ Top-$N_{\mathrm{keep}}$ clust.\ ($N_{\mathrm{keep}}\!=\!3$) & \color{gray} \cellcolor{secondcolor}26.2 & \cellcolor{secondcolor}8.9 & \cellcolor{secondcolor}13.2 & \cellcolor{secondcolor}11.0 & \cellcolor{secondcolor}11.0 & \color{gray} \cellcolor{secondcolor}30.6 & \cellcolor{secondcolor}16.8 & \cellcolor{secondcolor}18.5 & \cellcolor{secondcolor}16.3 & \cellcolor{secondcolor}17.2 \\
\;w/ Top-$N_{\mathrm{keep}}$ clust.\ ($N_{\mathrm{keep}}\!=\!5$) & \color{gray} 25.7 & 8.5 & 13.1 & 10.5 & 10.7 & \color{gray} 29.9 & 16.5 & 18.4 & 15.6 & 16.8 \\
\;w/o Clustering \vphantom{($N_{\mathrm{keep}}\!=\!3$)} & \color{gray} 25.6 & 8.4 & 13.1 & 10.2 & 10.6 & \color{gray} 29.4 & 16.3 & 18.4 & 15.0 & 16.6 \\
\;w/ Hungarian assigner \vphantom{($N_{\mathrm{keep}}\!=\!3$)} & \color{gray} 20.0 & 7.9 & 8.6 & 8.8 & 8.4 & \color{gray} 25.6 & 16.3 & 13.9 & 13.9 & 14.7 \\
\bottomrule
\end{tabularx}
\end{table*}

\begin{figure}[t]
    \centering
    \input{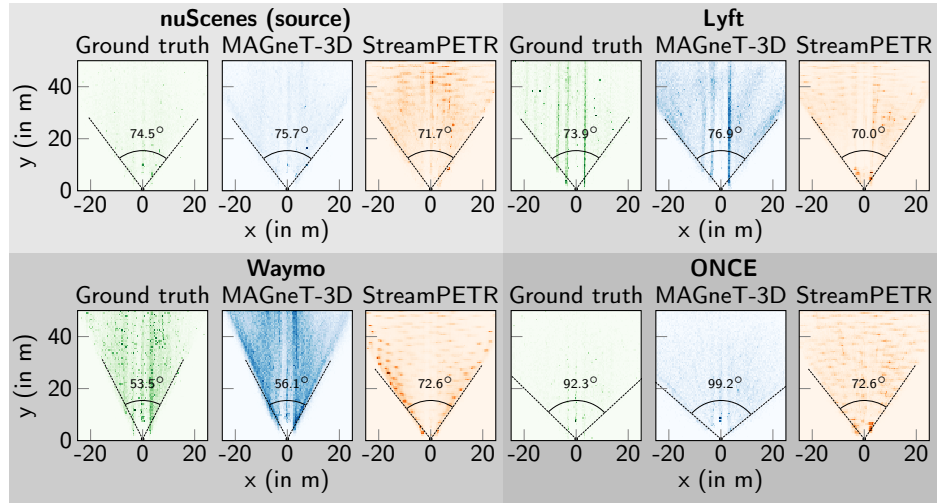}
    \caption{\textbf{Spatial distribution results.} Top-view distribution of 3D detections from \methodsymbol{} and our baseline StreamPETR~\cite{streampetr}. Both are trained on nuScenes~\cite{nuscene} (source dataset). The angles represent the prediction range (max-min azimuth). \methodsymbol{} adapts to different angles of the target datasets, while StreamPETR fails to adapt to the angle of the target datasets. More saturated colors indicate a higher density. Color coding: ground truth \colorindicator{gt1} (low density) $\xrightarrow{}$ \colorindicator{gt2} (high density); \methodsymbol{} \colorindicator{ours1} $\xrightarrow{}$ \colorindicator{ours2}; StreamPETR \colorindicator{spetr1} $\xrightarrow{}$ \colorindicator{spetr2}. Best viewed in color; zoom in for details.}
    \label{fig:prediction_range}
\end{figure}

\subsubsection{\propgationtransformersymbol{} Analysis.}
\Cref{tab:ablation_clustering} evaluates the impact of our dense training supervision and inference-time deduplication approach. Using a standard Hungarian assigner instead of our soft assigner reduces the average cross-dataset accuracy from \SI{16.6}{\%} to \SI{14.7}{\%} in NDS. This demonstrates that multi-hypothesis target supervision provides a richer, less brittle supervisory training signal for out-of-domain refinement. At inference time, our clustering step removes redundant 3D detections, improving average cross-dataset accuracy from \SI{16.6}{\%} to \SI{18.6}{\%} in NDS. Restricting each cluster to a tight threshold of $N_{\mathrm{keep}}=1$ yields the best accuracy. In particular, for $N_{\mathrm{keep}} = 5$ average cross-dataset accuracy drops from \SI{18.6}{\%} ($N_{\mathrm{keep}} = 1$) to \SI{16.8}{\%} in NDS. This demonstrates that retaining additional predictions ($N_{\mathrm{keep}} > 1$) reintroduces near-duplicate 3D boxes that compete during evaluation, consistently reducing accuracy across the source dataset and all target datasets.

\begin{figure}[t]
    \centering
    \input{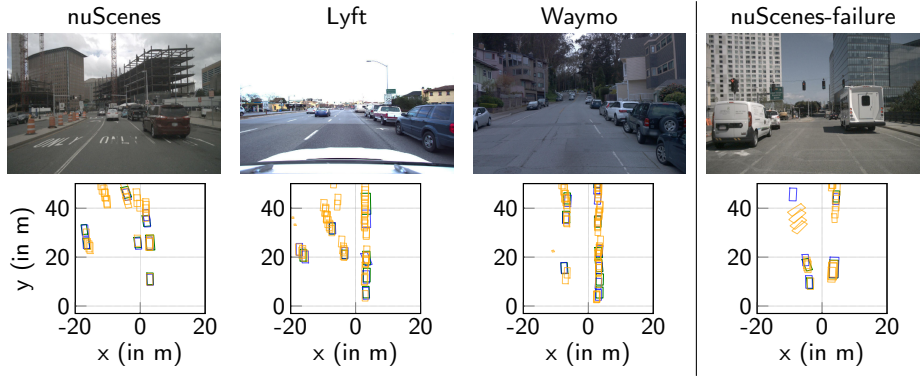}
    \caption{\textbf{Qualitative results.} \methodsymbol{} yields fewer false positives and more accurate predictions than StreamPETR \cite{streampetr}. All models are trained on nuScenes \cite{nuscene} and evaluated across nuScenes, Lyft \cite{lyft}, and Waymo \cite{waymo}. Best viewed in color; zoom in for details. BEV color coding: Ground truth \colorindicator{colorgt}, StreamPETR \colorindicator{colorbaseline}, and MAGneT-3D \colorindicator{colorours}.}
    \label{fig:qualitative_results}
\end{figure}

\subsection{Prediction Distribution Analysis}
\Cref{fig:prediction_range} visualizes the spatial distribution of predicted 3D bounding box centers for StreamPETR \cite{streampetr} and \methodsymbol{} when trained on nuScenes \cite{nuscene}. Due to its static queries, StreamPETR's predictions remain strictly confined within the source training distribution (\ie, nuScenes) across all target datasets. This creates severe local blind spots when encountering target spatial distribution shifts, failing to expand on wider settings (\eg, on ONCE \cite{once}), and remains overextended on smaller settings (\eg, on Waymo \cite{waymo}). Conversely, \methodsymbol{} generalizes and adapts to the spatial distribution of the target datasets and closely mirrors the true ground-truth distribution. In particular, while trained on nuScenes (\ang{74.5}), \methodsymbol{} expresses an angle of \ang{99.2} on ONCE with a ground-truth angle of \ang{92.3}, while StreamPETR only captures an angle of \ang{72.6}. This demonstrates that our dynamically derived queries do not suffer from source dataset biases and generalize to different target spatial distributions.

\subsection{Qualitative Results}
\Cref{fig:qualitative_results} provides qualitative detection results of our \methodsymbol{} and StreamPETR \cite{streampetr}.
\methodsymbol{} predicts more accurate box positions across diverse conditions and illumination levels. \methodsymbol{} dramatically reduces false-posi\-tive detections across all domains, demonstrating the practical effectiveness of our inference-time clustering step. The \emph{rightmost} column of \cref{fig:qualitative_results} shows an in-domain failure case, where our model generates a low-quality bounding box for a partially occluded vehicle and underestimates the depth of a far-range target.

\section{Conclusion}
\label{sec:conclusion}

We showed that state-of-the-art query-based 3D object detectors suffer from source-dataset biases. Predicted 3D detections are trapped within the source dataset's spatial distribution. To overcome this, we proposed \methodsymbol{}, a temporal monocular 3D detector composed of two novel components. \emph{First}, \anchorgensymbol{} constructs adaptive and dynamic per-frame detection proposals from image features. \emph{Second}, \propgationtransformersymbol{} provides dense and rich supervision during training and effectively filters duplicate 3D detection predictions during inference. We introduced a new cross-dataset benchmark to enable evaluation of monocular temporal 3D detectors under zero-shot domain shifts. On this cross-dataset benchmark, \methodsymbol{} establishes a new state of the art for domain generalization by a significant margin, while also improving in-domain accuracy.

{\small \subsubsection{Acknowledgments.} This work was supported by the ERC Advanced Grant SIMULACRON, the Georg Nemetschek Institute project AI4TWINNING, and the DFG project 4D-YouTube CR 250/26-1. Christoph Reich is supported by the Konrad Zuse School of Excellence in Learning and Intelligent Systems (\href{https://eliza.school}{ELIZA}) through the DAAD programme Konrad Zuse Schools of Excellence in Artificial Intelligence, sponsored by the Federal Ministry of Education and Research. Finally, we also acknowledge the support of the European Laboratory for Learning and Intelligent Systems (ELLIS).}

\bibliographystyle{splncs04}
\bibliography{main}

\begin{thebibliography}{10}
\providecommand{\url}[1]{\texttt{#1}}
\providecommand{\urlprefix}{URL }
\providecommand{\doi}[1]{https://doi.org/#1}

\bibitem{transfusion}
Bai, X., Hu, Z., Zhu, X., Huang, Q., Chen, Y., Fu, H., Tai, C.L.:
  {TransFusion}: {R}obust {LiDAR}-camera fusion for {3D} object detection with
  transformers. In: CVPR. pp. 1090--1099 (2022).
  \doi{10.1109/CVPR52688.2022.00116}

\bibitem{omni3d}
Brazil, G., Kumar, A., Straub, J., Ravi, N., Johnson, J., Gkioxari, G.:
  {Omni3D}: {A} large benchmark and model for {3D} object detection in the
  wild. In: CVPR. pp. 13154--13164 (2023). \doi{10.1109/CVPR52729.2023.01264}

\bibitem{kinematic}
Brazil, G., Pons-Moll, G., Liu, X., Schiele, B.: Kinematic {3D} object
  detection in monocular video. In: ECCV. vol. 12368, pp. 135--152 (2020).
  \doi{10.1007/978-3-030-58592-1_9}

\bibitem{nuscene}
Caesar, H., Bankiti, V., Lang, A.H., Vora, S., Liong, V.E., Xu, Q., Krishnan,
  A., Pan, Y., Baldan, G., Beijbom, O.: {nuScenes}: {A} multimodal dataset for
  autonomous driving. In: CVPR. pp. 11621--11631 (2020).
  \doi{10.1109/CVPR42600.2020.01164}

\bibitem{unified_mv_dg}
Chang, G., Lee, J., Kim, D., Kim, J., Lee, D., Ji, D., Jang, S., Kim, S.:
  Unified domain generalization and adaptation for multi-view {3D} object
  detection. In: NeurIPS. vol.~37, pp. 58498--58524 (2024),
  \url{https://proceedings.neurips.cc/paper_files/paper/2024/file/6b7e1e96243c9edc378f85e7d232e415-Paper-Conference.pdf}

\bibitem{monotdf}
Chen, X.Z., Chiu, Y.K., Huang, C.S., Chen, Y.L.: {MonoTDF}: {T}emporal deep
  feature learning for generalizable monocular {3D} object detection. Pattern
  Recognit.  \textbf{176},  113184 (2026). \doi{10.1016/j.patcog.2026.113184}

\bibitem{Voxelnext}
Chen, Y., Liu, J., Zhang, X., Qi, X., Jia, J.: {VoxelNeXt}: {F}ully sparse
  {VoxelNet} for {3D} object detection and tracking. In: CVPR. pp. 21674--21683
  (2023). \doi{10.1109/CVPR52729.2023.02076}

\bibitem{imagenet}
Deng, J., Dong, W., Socher, R., Li, L.J., Li, K., Fei-Fei, L.: {ImageNet}: {A}
  large-scale hierarchical image database. In: CVPR. pp. 248--255 (2009).
  \doi{10.1109/CVPR.2009.5206848}

\bibitem{mssvt}
Dong, S., Ding, L., Wang, H., Xu, T., Xu, X., Wang, J., Bian, Z., Wang, Y., Li,
  J.: {MsSVT}: {M}ixed-scale sparse voxel transformer for {3D} object detection
  on point clouds. In: NeurIPS. vol.~35, pp. 11615--11628 (2022),
  \url{https://proceedings.neurips.cc/paper_files/paper/2022/file/4bad7c27534efca029ca0d366c47c0e3-Paper-Conference.pdf}

\bibitem{lidar_generalization_study}
Eskandar, G.: An empirical study of the generalization ability of {Lidar} {3D}
  object detectors to unseen domains. In: CVPR. pp. 23815--23825 (2024).
  \doi{10.1109/CVPR52733.2024.02248}

\bibitem{domainbed}
Gulrajani, I., Lopez-Paz, D.: In search of lost domain generalization. In: ICLR
  (2021), \url{https://openreview.net/pdf?id=lQdXeXDoWtI}

\bibitem{resnet}
He, K., Zhang, X., Ren, S., Sun, J.: Deep residual learning for image
  recognition. In: CVPR. pp. 770--778 (2016). \doi{10.1109/CVPR.2016.90}

\bibitem{lyft}
Houston, J., Zuidhof, G., Bergamini, L., Ye, Y., Chen, L., Jain, A., Omari, S.,
  Iglovikov, V., Ondruska, P.: One thousand and one hours: {S}elf-driving
  motion prediction dataset. In: CoRL. vol.~155, pp. 409--418 (2021),
  \url{https://proceedings.mlr.press/v155/houston21a/houston21a.pdf}

\bibitem{quasimono}
Hu, H.N., Yang, Y.H., Fischer, T., Darrell, T., Yu, F., Sun, M.: Monocular
  quasi-dense {3D} object tracking. IEEE Trans. Pattern Anal. Mach. Intell.
  \textbf{45}(2),  1992--2008 (2022). \doi{10.1109/TPAMI.2022.3168781}

\bibitem{bevdet4d}
Huang, J., Huang, G.: {BEVDet4D}: {E}xploit temporal cues in multi-camera {3D}
  object detection. arXiv:2203.17054 [cs.CV]  (2022).
  \doi{10.48550/arXiv.2203.17054}

\bibitem{bevdet}
Huang, J., Huang, G., Zhu, Z., Ye, Y., Du, D.: {BEVDet}: {H}igh-performance
  multi-camera {3D} object detection in bird-eye-view. arXiv:2112.11790 [cs.CV]
   (2021). \doi{10.48550/arXiv.2112.11790}

\bibitem{gate3d}
Im, E., Jee, C., Lee, J.K.: {GATE3D}: {G}eneralized attention-based
  task-synergized estimation in {3D}. arXiv:2504.11014 [cs.CV]  (2025).
  \doi{10.48550/arXiv.2504.11014}

\bibitem{monouni}
Jia, J., Li, Z., Shi, Y.: {MonoUNI}: {A} unified vehicle and
  infrastructure-side monocular {3D} object detection network with sufficient
  depth clues. In: NeurIPS. vol.~36, pp. 11703--11715 (2023),
  \url{https://proceedings.neurips.cc/paper_files/paper/2023/file/2703a0e3c2b33506295a77762338cf24-Paper-Conference.pdf}

\bibitem{far3d}
Jiang, X., Li, S., Liu, Y., Wang, S., Jia, F., Wang, T., Han, L., Zhang, X.:
  {Far3D}: {E}xpanding the horizon for surround-view {3D} object detection. In:
  AAAI. vol.~38, pp. 2561--2569 (2024). \doi{10.1609/aaai.v38i3.28033}

\bibitem{supcon}
Khosla, P., Teterwak, P., Wang, C., Sarna, A., Tian, Y., Isola, P., Maschinot,
  A., Liu, C., Krishnan, D.: Supervised contrastive learning. In: NeurIPS.
  vol.~33, pp. 18661--18673 (2020),
  \url{https://proceedings.neurips.cc/paper/2020/file/d89a66c7c80a29b1bdbab0f2a1a94af8-Paper.pdf}

\bibitem{tsmabev}
Kong, L., Xie, S., Hu, H., Niu, Y., Ooi, W.T., Cottereau, B.R., Ng, L.X., Ma,
  Y., Zhang, W., Pan, L., et~al.: The {RoboDrive} challenge: {D}rive anytime
  anywhere in any condition. arXiv:2405.08816 [cs.CV]  (2024).
  \doi{10.48550/arXiv.2405.08816}

\bibitem{3dvfield}
Lehner, A., Gasperini, S., Marcos-Ramiro, A., Schmidt, M., Mahani, M.A.N.,
  Navab, N., Busam, B., Tombari, F.: {3D-VField}: {A}dversarial augmentation of
  point clouds for domain generalization in {3D} object detection. In: CVPR.
  pp. 17295--17304 (2022). \doi{10.1109/CVPR52688.2022.01678}

\bibitem{deepfusion}
Li, Y., Yu, A.W., Meng, T., Caine, B., Ngiam, J., Peng, D., Shen, J., Lu, Y.,
  Zhou, D., Le, Q.V., Yuille, A., Tan, M.: {DeepFusion}: {L}idar-camera deep
  fusion for multi-modal {3D} object detection. In: CVPR. pp. 17182--17191
  (2022). \doi{10.1109/CVPR52688.2022.01667}

\bibitem{monolss}
Li, Z., Jia, J., Shi, Y.: {MonoLSS}: {L}earnable sample selection for monocular
  {3D} detection. In: 3DV. pp. 1125--1135 (2024).
  \doi{10.1109/3DV62453.2024.00088}

\bibitem{dgmono3d}
Li, Z., Chen, Z., Li, A., Fang, L., Jiang, Q., Liu, X., Jiang, J.: Towards
  model generalization for monocular {3D} object detection. arXiv:2205.11664
  [cs.CV]  (2022). \doi{10.48550/arXiv.2205.11664}

\bibitem{stmono3d}
Li, Z., Chen, Z., Li, A., Fang, L., Jiang, Q., Liu, X., Jiang, J.: Unsupervised
  domain adaptation for monocular {3D} object detection via self-training. In:
  ECCV. vol. 13669, pp. 245--262 (2022). \doi{10.1007/978-3-031-20077-9_15}

\bibitem{bevformer}
Li, Z., Wang, W., Li, H., Xie, E., Sima, C., Lu, T., Yu, Q., Dai, J.:
  {BEVFormer}: {L}earning bird's-eye-view representation from {LiDAR}-camera
  via spatiotemporal transformers. IEEE Trans. Pattern Anal. Mach. Intell.
  \textbf{47}(3),  2020--2036 (2024). \doi{10.1109/TPAMI.2024.3515454}

\bibitem{bevfusion}
Liang, T., Xie, H., Yu, K., Xia, Z., Lin, Z., Wang, Y., Tang, T., Wang, B.,
  Tang, Z.: {BEVFusion}: {A} simple and robust {LiDAR}-camera fusion framework.
  In: NeurIPS. vol.~35, pp. 10421--10434 (2022),
  \url{https://proceedings.neurips.cc/paper_files/paper/2022/file/43d2b7fbee8431f7cef0d0afed51c691-Paper-Conference.pdf}

\bibitem{monotta}
Lin, H., Zhang, Y., Niu, S., Cui, S., Li, Z.: {MonoTTA}: {F}ully test-time
  adaptation for monocular {3D} object detection. In: ECCV. vol. 15102, pp.
  96--114 (2024). \doi{10.1007/978-3-031-72784-9_6}

\bibitem{sparse4d}
Lin, X., Lin, T., Pei, Z., Huang, L., Su, Z.: {Sparse4D}: {M}ulti-view {3D}
  object detection with sparse spatial-temporal fusion. arXiv:2211.10581
  [cs.CV]  (2022). \doi{10.48550/arXiv.2211.10581}

\bibitem{sparse4dv2}
Lin, X., Lin, T., Pei, Z., Huang, L., Su, Z.: {Sparse4D v2}: {R}ecurrent
  temporal fusion with sparse model. arXiv:2305.14018 [cs.CV]  (2023).
  \doi{10.48550/arXiv.2305.14018}

\bibitem{sparse4dv3}
Lin, X., Pei, Z., Lin, T., Huang, L., Su, Z.: {Sparse4D} v3: Advancing
  end-to-end {3D} detection and tracking. arXiv:2311.11722 [cs.CV]  (2023).
  \doi{10.48550/arXiv.2311.11722}

\bibitem{rcbevdet}
Lin, Z., Liu, Z., Xia, Z., Wang, X., Wang, Y., Qi, S., Dong, Y., Dong, N.,
  Zhang, L., Zhu, C.: {RCBEVDet}: {R}adar-camera fusion in bird's eye view for
  {3D} object detection. In: CVPR. pp. 14928--14937 (2024).
  \doi{10.1109/CVPR52733.2024.01414}

\bibitem{monotakd}
Liu, H., Wu, C., Cheng, J., Chai, W., Wang, S., Liu, G., Latapie, H., Wu, J.,
  Hwang, J., Shuai, H., Cheng, W.: {MonoTAKD}: {T}eaching assistant knowledge
  distillation for monocular {3D} object detection. In: CVPR. pp. 22266--22275
  (2025). \doi{10.1109/CVPR52734.2025.02074}

\bibitem{petr}
Liu, Y., Wang, T., Zhang, X., Sun, J.: {PETR}: {P}osition embedding
  transformation for multi-view {3D} object detection. In: ECCV. vol. 13687,
  pp. 531--548 (2022). \doi{10.1007/978-3-031-19812-0_31}

\bibitem{petrv2}
Liu, Y., Yan, J., Jia, F., Li, S., Gao, A., Wang, T., Zhang, X.: {PETRv2}: {A}
  unified framework for {3D} perception from multi-camera images. In: ICCV. pp.
  3262--3272 (2023). \doi{10.1109/ICCV51070.2023.00302}

\bibitem{smoke}
Liu, Z., Wu, Z., T{\'o}th, R.: {SMOKE}: {S}ingle-stage monocular {3D} object
  detection via keypoint estimation. In: CVPRW. pp. 996--997 (2020).
  \doi{10.1109/CVPRW50498.2020.00506}

\bibitem{adamw}
Loshchilov, I., Hutter, F.: Decoupled weight decay regularization. In: ICLR
  (2019), \url{https://openreview.net/pdf?id=Bkg6RiCqY7}

\bibitem{lsf}
Lu, H.C., Lin, C.Y., Hsu, W.H.: Improving generalization ability for {3D}
  object detection by learning sparsity-invariant features. In: ICRA. pp.
  10892--10898 (2025). \doi{10.1109/ICRA55743.2025.11128274}

\bibitem{once}
Mao, J., Niu, M., Jiang, C., Liang, H., Chen, J., Liang, X., Li, Y., Ye, C.,
  Zhang, W., Li, Z., Yu, J., Xu, H., Xu, C.: One million scenes for autonomous
  driving: {ONCE} dataset. In: NeurIPS D\&B (2021),
  \url{https://datasets-benchmarks-proceedings.neurips.cc/paper/2021/file/67c6a1e7ce56d3d6fa748ab6d9af3fd7-Paper-round1.pdf}

\bibitem{ideal-m3d}
Meier, J., G{\"{u}}nther, F., Marin, R., Dhaouadi, O., Kaiser, J., Cremers, D.:
  {IDEAL-M3D}: {I}nstance diversity-enriched active learning for monocular {3D}
  detection. In: WACV. pp. 181--191 (2026). \doi{10.1109/WACV61042.2026.00026}

\bibitem{monoct}
Meier, J., Inchingolo, L., Dhaouadi, O., Xia, Y., Kaiser, J., Cremers, D.:
  {MonoCT}: {O}vercoming monocular {3D} detection domain shift with consistent
  teacher models. In: ICRA. pp. 351--358 (2025).
  \doi{10.1109/ICRA55743.2025.11127874}

\bibitem{meier2025lead}
Meier, J., Michel, J., Dhaouadi, O., Yang, Y.H., Reich, C., Bauer, Z., Roth,
  S., Pollefeys, M., Kaiser, J., Cremers, D.: {LeAD-M3D}: {L}everaging
  asymmetric distillation for real-time monocular {3D} detection.
  arXiv:2512.05663 [cs.CV]  (2025). \doi{10.48550/arXiv.2512.05663}

\bibitem{carladrone}
Meier, J., Scalerandi, L., Dhaouadi, O., Kaiser, J., Araslanov, N., Cremers,
  D.: {CARLA Drone}: {M}onocular {3D} object detection from a different
  perspective. In: GCPR. vol. 15298, pp. 137--152 (2024).
  \doi{10.1007/978-3-031-85187-2_9}

\bibitem{centerfusion}
Nabati, R., Qi, H.: {CenterFusion}: Center-based radar and camera fusion for
  {3D} object detection. In: WACV. pp. 1527--1536 (2021).
  \doi{10.1109/WACV48630.2021.00157}

\bibitem{transcar}
Pang, S., Morris, D., Radha, H.: {TransCAR}: {T}ransformer-based
  camera-and-radar fusion for {3D} object detection. In: IROS. pp. 10902--10909
  (2023). \doi{10.1109/IROS55552.2023.10341793}

\bibitem{occupancym3d}
Peng, L., Xu, J., Cheng, H., Yang, Z., Wu, X., Qian, W., Wang, W., Wu, B., Cai,
  D.: Learning occupancy for monocular {3D} object detection. In: CVPR. pp.
  10281--10292 (2024). \doi{10.1109/CVPR52733.2024.00979}

\bibitem{lss}
Philion, J., Fidler, S.: {Lift, Splat, Shoot}: {E}ncoding images from arbitrary
  camera rigs by implicitly unprojecting to {3D}. In: ECCV. vol. 12359, pp.
  194--210 (2020). \doi{10.1007/978-3-030-58568-6_12}

\bibitem{monodgp}
Pu, F., Wang, Y., Deng, J., Yang, W.: {MonoDGP}: {M}onocular {3D} object
  detection with decoupled-query and geometry-error priors. In: CVPR. pp.
  6520--6530 (2025). \doi{10.1109/CVPR52734.2025.00611}

\bibitem{mondiff}
Ranasinghe, Y., Hegde, D., Patel, V.M.: {MonoDiff}: {M}onocular {3D} object
  detection and pose estimation with diffusion models. In: CVPR. pp.
  10659--10670 (2024). \doi{10.1109/CVPR52733.2024.01014}

\bibitem{cadnn}
Reading, C., Harakeh, A., Chae, J., Waslander, S.L.: Categorical depth
  distribution network for monocular {3D} object detection. In: CVPR. pp.
  8555--8564 (2021). \doi{10.1109/CVPR46437.2021.00845}

\bibitem{oft}
Roddick, T., Kendall, A., Cipolla, R.: Orthographic feature transform for
  monocular {3D} object detection. In: BMVC (2019)

\bibitem{waymo}
Sun, P., Kretzschmar, H., Dotiwalla, X., Chouard, A., Patnaik, V., Tsui, P.,
  Guo, J., Zhou, Y., Chai, Y., Caine, B., et~al.: Scalability in perception for
  autonomous driving: {W}aymo open dataset. In: CVPR. pp. 2446--2454 (2020).
  \doi{10.1109/CVPR42600.2020.00252}

\bibitem{robotic2}
Tanveer, M.H., Fatima, Z., Mariam, H., Rehman, T., Voicu, R.C.:
  Three-dimensional outdoor object detection in quadrupedal robots for
  surveillance navigations. Actuators  \textbf{13}(10) (2024).
  \doi{10.3390/act13100422}

\bibitem{clipthegap}
Vidit, V., Engilberge, M., Salzmann, M.: {CLIP} the gap: {A} single domain
  generalization approach for object detection. In: CVPR. pp. 3219--3229
  (2023). \doi{10.1109/CVPR52729.2023.00314}

\bibitem{voxelrcnn}
Wang, H., Chen, Z., Cai, Y., Chen, L., Li, Y., Sotelo, M.A., Li, Z.:
  {Voxel-RCNN-complex}: {A}n effective {3D} point cloud object detector for
  complex traffic conditions. IEEE Trans. Instrum. Meas.  \textbf{71},  1--12
  (2022). \doi{10.1109/TIM.2022.3165251}

\bibitem{robotic1}
Wang, L., Li, R., Sun, J., Liu, X., Zhao, L., Seah, H.S., Quah, C.K.,
  Tandianus, B.: Multi-view fusion-based {3D} object detection for robot indoor
  scene perception. Sensors  \textbf{19}(19), ~4092 (2019).
  \doi{10.3390/s19194092}

\bibitem{focalpetr}
Wang, S., Jiang, X., Li, Y.: {Focal-PETR}: {E}mbracing foreground for efficient
  multi-camera {3D} object detection. IEEE Trans. Intell. Veh.  \textbf{9}(1),
  1481--1489 (2024). \doi{10.1109/TIV.2023.3332608}

\bibitem{streampetr}
Wang, S., Liu, Y., Wang, T., Li, Y., Zhang, X.: Exploring object-centric
  temporal modeling for efficient multi-view {3D} object detection. In: ICCV.
  pp. 3621--3631 (2023). \doi{10.1109/ICCV51070.2023.00335}

\bibitem{dgbev}
Wang, S., Zhao, X., Xu, H.M., Chen, Z., Yu, D., Chang, J., Yang, Z., Zhao, F.:
  Towards domain generalization for multi-view {3D} object detection in
  bird-eye-view. In: CVPR. pp. 13333--13342 (2023).
  \doi{10.1109/CVPR52729.2023.01281}

\bibitem{fcos3d}
Wang, T., Zhu, X., Pang, J., Lin, D.: {FCOS3D}: {F}ully convolutional one-stage
  monocular {3D} object detection. In: ICCV. pp. 913--922 (2021).
  \doi{10.1109/ICCVW54120.2021.00107}

\bibitem{detr3d}
Wang, Y., Guizilini, V.C., Zhang, T., Wang, Y., Zhao, H., Solomon, J.:
  {DETR3D}: {3D} object detection from multi-view images via {3D}-to-{2D}
  queries. In: CoRL. vol.~164, pp. 180--191 (2022),
  \url{https://proceedings.mlr.press/v164/wang22b/wang22b.pdf}

\bibitem{monocd}
Yan, L., Yan, P., Xiong, S., Xiang, X., Tan, Y.: {MonoCD}: {M}onocular {3D}
  object detection with complementary depths. In: CVPR. pp. 10248--10257
  (2024). \doi{10.1109/CVPR52733.2024.00976}

\bibitem{bevformerv2}
Yang, C., Chen, Y., Tian, H., Tao, C., Zhu, X., Zhang, Z., Huang, G., Li, H.,
  Qiao, Y., Lu, L., et~al.: {BEVFormer v2}: {A}dapting modern image backbones
  to bird's-eye-view recognition via perspective supervision. In: CVPR. pp.
  17830--17839 (2023). \doi{10.1109/CVPR52729.2023.01710}

\bibitem{MonoGDG}
Yang, F., Chen, H., He, Y., Zhao, S., Zhang, C., Ni, K., Ding, G.:
  Geometry-guided domain generalization for monocular {3D} object detection.
  In: AAAI. vol.~38, pp. 6467--6476 (2024). \doi{10.1609/aaai.v38i6.28467}

\bibitem{dbqssd}
Yang, J., Song, L., Liu, S., Mao, W., Li, Z., Li, X., Sun, H., Sun, J., Zheng,
  N.: {DBQ-SSD}: {D}ynamic ball query for efficient {3D} object detection. In:
  ICLR (2023), \url{https://openreview.net/pdf?id=ZccFLU-Yk65}

\bibitem{3dssd}
Yang, Z., Sun, Y., Liu, S., Jia, J.: {3DSSD}: {P}oint-based {3D} single stage
  object detector. In: CVPR. pp. 11040--11048 (2020).
  \doi{10.1109/CVPR42600.2020.01105}

\bibitem{Std:Sparse-to-dense}
Yang, Z., Sun, Y., Liu, S., Shen, X., Jia, J.: {STD}: {S}parse-to-dense {3D}
  object detector for point cloud. In: ICCV. pp. 1951--1960 (2019).
  \doi{10.1109/ICCV.2019.00204}

\bibitem{rope3d}
Ye, X., Shu, M., Li, H., Shi, Y., Li, Y., Wang, G., Tan, X., Ding, E.:
  {Rope3D}: {T}he roadside perception dataset for autonomous driving and
  monocular {3D} object detection task. In: CVPR. pp. 21309--21318 (2022).
  \doi{10.1109/CVPR52688.2022.02065}

\bibitem{objectaspoints}
Zhou, X., Wang, D., Kr{\"a}henb{\"u}hl, P.: Objects as points. arXiv:1904.07850
  [cs.CV]  (2019). \doi{10.48550/arXiv.1904.07850}

\bibitem{mogde}
Zhou, Y., Liu, Q., Zhu, H., Li, Y., Chang, S., Guo, M.: {MoGDE}: {B}oosting
  mobile monocular {3D} object detection with ground depth estimation. In:
  NeurIPS. vol.~35, pp. 2033--2045 (2022),
  \url{https://papers.neurips.cc/paper_files/paper/2022/file/0d81e6f2511fc78631ee0315fafeef9e-Paper-Conference.pdf}

\bibitem{monoatt}
Zhou, Y., Zhu, H., Liu, Q., Chang, S., Guo, M.: {MonoATT}: {O}nline monocular
  {3D} object detection with adaptive token transformer. In: CVPR. pp.
  17493--17503 (2023). \doi{10.1109/CVPR52729.2023.01678}

\end{thebibliography}
\end{document}